\documentclass[10pt,a4paper]{article}
\usepackage[T1]{fontenc}
\usepackage{newtxtext,newtxmath}
\usepackage[left=20.5mm,right=20.5mm,top=25mm,bottom=25mm,headheight=24pt,headsep=13pt]{geometry}
\usepackage{amsmath,mathtools,booktabs,tabularx,array,multirow}
\usepackage{graphicx,xcolor,hyperref,caption,enumitem,microtype,titlesec,placeins,needspace,fancyhdr}
\usepackage{algorithm,algpseudocode}
\definecolor{Blue}{HTML}{2463B4}
\definecolor{YANblue}{HTML}{2463B4}
\hypersetup{colorlinks=true,linkcolor=Blue,citecolor=black,urlcolor=Blue,
 pdftitle={EdgeRock: Strong Reasoning at Constant Memory},
 pdfauthor={Huishan Ji, Weiming Zhang, Qirui Ye, Jinchao Xu, Yike Guo, Hua Xu}}
\setlist{nosep,leftmargin=1.3em}
\renewcommand{\arraystretch}{1.12}
\newcolumntype{Y}{>{\raggedright\arraybackslash}X}
\newcolumntype{C}[1]{>{\centering\arraybackslash}p{#1}}
\titleformat{\section}{\large\bfseries}{\thesection}{.65em}{}
\titleformat{\subsection}{\normalsize\bfseries}{\thesubsection}{.65em}{}
\titleformat{\subsubsection}{\normalsize\itshape}{\thesubsubsection}{.65em}{}
\titlespacing*{\section}{0pt}{12pt plus 2pt minus 2pt}{6pt}
\titlespacing*{\subsection}{0pt}{9pt plus 2pt minus 2pt}{4pt}
\titlespacing*{\subsubsection}{0pt}{7pt}{3pt}
\graphicspath{{figures/}}

\newcommand{\CoreReasoningResults}{%
\begin{table}[!htbp]
\centering
\caption{Competition mathematics and MATH-500 (mean pass@1, \%). Retained gives EdgeRock-4B as a percentage of its source model, Qwen3.5-4B. AIME averages 2024--2026.}
\label{tab:core-reasoning}
\small
\begin{tabularx}{\textwidth}{@{}Y r r r@{}}
\toprule
Benchmark & EdgeRock-4B & Qwen3.5-4B~\cite{qwen35} & Retained (\%) \\
\midrule
AIME 2024~\cite{aime} & 86.7 & 89.6 & 96.8 \\
AIME 2025~\cite{aime} & 77.4 & 80.8 & 95.8 \\
AIME 2026~\cite{aime} & 84.6 & 87.9 & 96.2 \\
HMMT Feb 2025~\cite{hmmt} & 71.8 & 74.6 & 96.2 \\
HMMT Nov 2025~\cite{hmmt} & 69.8 & 78.3 & 89.1 \\
HMMT Feb 2026~\cite{hmmt} & 63.6 & 65.9 & 96.5 \\
MATH-500~\cite{mathds,math500} & 97.5 & 98.0 & 99.5 \\
\midrule
AIME three-year mean & 82.9 & 86.1 & 96.3 \\
\bottomrule
\end{tabularx}
\end{table}
}

\newcommand{\BoundedMathResults}{%
\begin{table}[!htbp]
\centering
\caption{Bounded-state mathematics scores (\%). MATH: MATH-500.}
\label{tab:all-math}
\fontsize{8.7}{10.8}\selectfont
\setlength{\tabcolsep}{3pt}
\renewcommand{\arraystretch}{1.12}
\begin{tabularx}{\textwidth}{@{}Y r r r r r r r r @{}}
\toprule
Model & \shortstack[r]{GSM8K\\\cite{gsm8k}} & \shortstack[r]{MATH\\\cite{math500}} & \shortstack[r]{AIME '24\\\cite{aime}} & \shortstack[r]{AIME '25\\\cite{aime}} & \shortstack[r]{AIME '26\\\cite{aime}} & \shortstack[r]{HMMT\\Feb '25\\\cite{hmmt}} & \shortstack[r]{HMMT\\Nov '25\\\cite{hmmt}} & \shortstack[r]{HMMT\\Feb '26\\\cite{hmmt}} \\
\midrule
\textcolor{YANblue}{\textbf{EdgeRock-4B}} & 94.5 & \textbf{97.5} & \textbf{86.7} & \textbf{77.4} & \textbf{84.6} & \textbf{71.8} & \textbf{69.8} & \textbf{63.6} \\
ARWKV-R1 1.5B~\cite{arwkv} & 24.8 & 13.6 & 0.0 & 0.0 & 0.0 & 0.4 & 0.8 & 1.1 \\
RWKV-7 G1i 1.5B~\cite{rwkvg1} & 60.3 & 28.2 & 0.0 & 0.4 & 0.4 & 1.7 & 5.0 & 0.0 \\
RWKV-7 G1j 1.5B~\cite{rwkvg1} & 57.2 & 31.4 & 1.2 & 2.1 & 0.0 & 1.7 & 4.2 & 0.0 \\
RWKV-7 G1i 2.9B~\cite{rwkvg1} & 71.7 & 43.8 & 1.7 & 1.7 & 2.1 & 1.7 & 5.0 & 0.8 \\
RWKV-7 G1j 2.9B~\cite{rwkvg1} & 74.1 & 48.2 & 0.4 & 4.6 & 1.7 & 2.1 & 5.0 & 0.4 \\
AHN-GDN Qwen2.5 3B~\cite{ahn} & 83.8 & 66.2 & 7.1 & 2.1 & 3.3 & 2.9 & 6.7 & 4.9 \\
ARWKV-R1 7B~\cite{arwkv} & 51.2 & 33.2 & 0.0 & 0.0 & 0.4 & 0.8 & 1.7 & 1.1 \\
Falcon3-Mamba 7B Instruct~\cite{falcon3} & 61.4 & 41.8 & 1.7 & 0.8 & 1.2 & 0.8 & 5.4 & 2.7 \\
QRWKV7 7B Instruct~\cite{radlads} & 65.5 & 41.0 & 0.0 & 0.0 & 0.0 & 0.0 & 6.7 & 0.0 \\
xLSTM 7B~\cite{xlstm7b} & 43.4 & 25.0 & 0.0 & 0.0 & 0.0 & 1.7 & 5.0 & 1.5 \\
RWKV-7 G1i 7.2B~\cite{rwkvg1} & 85.1 & 68.6 & 8.8 & 11.7 & 7.9 & 2.1 & 5.0 & 3.4 \\
RWKV-7 G1j 7.2B~\cite{rwkvg1} & 81.8 & 64.2 & 8.3 & 11.2 & 4.6 & 4.2 & 4.2 & 0.8 \\
RecurrentGemma 9B IT~\cite{recurrentgemma} & 51.7 & 22.8 & 0.4 & 0.8 & 0.0 & 1.2 & 5.4 & 1.9 \\
RWKV-7 G1j 13.3B~\cite{rwkvg1} & \textbf{94.6} & 77.2 & 17.5 & 27.5 & 11.2 & 3.3 & 5.8 & 12.1 \\
AHN-GDN Qwen2.5 14B~\cite{ahn} & 93.6 & 80.0 & 13.8 & 10.0 & 10.4 & 7.5 & 10.0 & 4.9 \\
Phi-3 Medium 4K Instruct~\cite{phi3window} & 85.3 & 49.0 & 3.3 & 3.3 & 0.0 & 6.7 & 2.5 & 3.0 \\
\bottomrule
\end{tabularx}
\end{table}
}

\newcommand{\GrowingMathResults}{%
\begin{table}[!htbp]
\centering
\caption{Mathematics comparison with growing-history models (\%). MATH: MATH-500.}
\label{tab:growing-math}
\fontsize{8.7}{10.8}\selectfont
\setlength{\tabcolsep}{3pt}
\renewcommand{\arraystretch}{1.12}
\begin{tabularx}{\textwidth}{@{}Y r r r r r r r r @{}}
\toprule
Model & \shortstack[r]{GSM8K\\\cite{gsm8k}} & \shortstack[r]{MATH\\\cite{math500}} & \shortstack[r]{AIME '24\\\cite{aime}} & \shortstack[r]{AIME '25\\\cite{aime}} & \shortstack[r]{AIME '26\\\cite{aime}} & \shortstack[r]{HMMT\\Feb '25\\\cite{hmmt}} & \shortstack[r]{HMMT\\Nov '25\\\cite{hmmt}} & \shortstack[r]{HMMT\\Feb '26\\\cite{hmmt}} \\
\midrule
\textcolor{YANblue}{\textbf{EdgeRock-4B}} & 94.5 & 97.5 & 86.7 & 77.4 & 84.6 & 71.8 & 69.8 & 63.6 \\
Qwen3.5-4B~\cite{qwen35} & \textbf{94.8} & \textbf{98.0} & \textbf{89.6} & 80.8 & 87.9 & 74.6 & 78.3 & 65.9 \\
Gemma4-E2B-it~\cite{gemma4} & 90.1 & 78.8 & 38.3 & 33.8 & 33.8 & 21.7 & 17.5 & 20.1 \\
MiniCPM5-2B~\cite{minicpm5} & 93.6 & 97.0 & 87.5 & 85.4 & 90.4 & 76.7 & 74.6 & 63.3 \\
LFM2.5-2.6B~\cite{lfm25config} & 91.0 & 92.7 & 63.4 & 56.3 & 48.9 & 34.7 & 42.9 & 32.9 \\
Qwen3.5-2B~\cite{qwen35} & 87.6 & 86.8 & 41.2 & 34.2 & 35.4 & 26.2 & 23.3 & 25.4 \\
Granite4.2-3B~\cite{granite42} & 93.8 & 97.8 & 85.0 & 78.8 & 82.9 & 75.8 & 72.5 & 59.5 \\
Phi-4-mini-reasoning~\cite{phi4mini} & 93.6 & 91.6 & 45.4 & 32.5 & 38.8 & 25.0 & 28.7 & 24.6 \\
Gemma4-E4B-it~\cite{gemma4} & 91.3 & 88.2 & 42.1 & 40.0 & 38.3 & 25.4 & 32.9 & 28.8 \\
Nemotron-3-Nano-4B~\cite{nemotron3nano} & 92.4 & 94.8 & 60.8 & 56.7 & 53.8 & 40.0 & 47.1 & 36.4 \\
Granite4.2-8B~\cite{granite42} & 94.4 & \textbf{98.0} & 88.3 & \textbf{87.9} & \textbf{91.7} & \textbf{81.7} & \textbf{86.7} & \textbf{73.9} \\
MiniCPM4.1-8B~\cite{minicpm41} & 93.6 & 96.4 & 77.1 & 70.0 & 70.8 & 58.3 & 59.6 & 49.2 \\
\bottomrule
\end{tabularx}
\end{table}
}

\newcommand{\BoundedKnowledgeResults}{%
\begin{table}[!htbp]
\centering
\caption{Bounded-state knowledge and science scores. Pro and Redux denote MMLU-Pro and MMLU-Redux; GPQA denotes GPQA-Diamond; Super denotes SuperGPQA.}
\label{tab:knowledge}
\fontsize{8.7}{10.8}\selectfont
\setlength{\tabcolsep}{3pt}
\renewcommand{\arraystretch}{1.12}
\begin{tabularx}{\textwidth}{@{}Y r r r r r r r @{}}
\toprule
Model & \shortstack[r]{MMLU\\\cite{mmlu}} & \shortstack[r]{Pro\\\cite{mmlupro}} & \shortstack[r]{Redux\\\cite{mmluredux}} & \shortstack[r]{CMMLU\\\cite{cmmlu}} & \shortstack[r]{C-Eval\\\cite{ceval}} & \shortstack[r]{GPQA\\\cite{gpqa}} & \shortstack[r]{Super\\\cite{supergpqa}} \\
\midrule
\textcolor{YANblue}{\textbf{EdgeRock-4B}} & \textbf{88.2} & \textbf{79.8} & \textbf{92.0} & 73.3 & \textbf{74.6} & \textbf{64.6} & \textbf{60.0} \\
ARWKV-R1 1.5B~\cite{arwkv} & 17.4 & 7.5 & 17.9 & 15.6 & 15.2 & 13.1 & 5.8 \\
RWKV-7 G1i 1.5B~\cite{rwkvg1} & 49.7 & 13.3 & 55.4 & 33.6 & 33.0 & 26.8 & 17.7 \\
RWKV-7 G1j 1.5B~\cite{rwkvg1} & 49.3 & 16.6 & 53.8 & 35.3 & 33.3 & 26.3 & 16.1 \\
RWKV-7 G1i 2.9B~\cite{rwkvg1} & 63.6 & 37.8 & 63.4 & 36.1 & 32.5 & 26.8 & 21.6 \\
RWKV-7 G1j 2.9B~\cite{rwkvg1} & 64.7 & 37.9 & 58.9 & 37.9 & 36.5 & 31.8 & 21.8 \\
AHN-GDN Qwen2.5 3B~\cite{ahn} & 62.7 & 42.8 & 65.7 & 62.4 & 62.9 & 31.3 & 23.0 \\
ARWKV-R1 7B~\cite{arwkv} & 32.3 & 8.8 & 35.2 & 29.0 & 29.0 & 13.1 & 8.8 \\
Falcon3-Mamba 7B Instruct~\cite{falcon3} & 57.7 & 26.1 & 64.8 & 33.4 & 35.2 & 20.7 & 12.2 \\
QRWKV7 7B Instruct~\cite{radlads} & 64.6 & 23.0 & 34.1 & 57.7 & 56.0 & 8.6 & 15.4 \\
xLSTM 7B~\cite{xlstm7b} & 19.6 & 4.6 & 37.9 & 16.9 & 19.5 & 12.1 & 5.3 \\
RWKV-7 G1i 7.2B~\cite{rwkvg1} & 69.4 & 44.9 & 73.5 & 49.8 & 51.6 & 30.3 & 25.7 \\
RWKV-7 G1j 7.2B~\cite{rwkvg1} & 67.5 & 46.2 & 72.7 & 50.8 & 48.9 & 36.9 & 25.3 \\
RecurrentGemma 9B IT~\cite{recurrentgemma} & 55.5 & 13.4 & 55.0 & 43.4 & 41.7 & 24.7 & 19.5 \\
RWKV-7 G1j 13.3B~\cite{rwkvg1} & 76.7 & 47.6 & 81.8 & 67.4 & 66.3 & 42.4 & 31.8 \\
AHN-GDN Qwen2.5 14B~\cite{ahn} & 77.2 & 63.5 & 83.5 & \textbf{75.0} & 74.2 & 40.9 & 35.5 \\
Phi-3 Medium 4K Instruct~\cite{phi3window} & 76.3 & 54.0 & 81.4 & 53.3 & 51.3 & 26.8 & 25.6 \\
\bottomrule
\end{tabularx}
\end{table}
}

\newcommand{\GrowingKnowledgeResults}{%
\begin{table}[!htbp]
\centering
\caption{Knowledge and science comparison with growing-history models. Pro and Redux denote MMLU-Pro and MMLU-Redux; GPQA denotes GPQA-Diamond; Super denotes SuperGPQA.}
\label{tab:growing-knowledge}
\fontsize{8.7}{10.8}\selectfont
\setlength{\tabcolsep}{3pt}
\renewcommand{\arraystretch}{1.12}
\begin{tabularx}{\textwidth}{@{}Y r r r r r r r @{}}
\toprule
Model & \shortstack[r]{MMLU\\\cite{mmlu}} & \shortstack[r]{Pro\\\cite{mmlupro}} & \shortstack[r]{Redux\\\cite{mmluredux}} & \shortstack[r]{CMMLU\\\cite{cmmlu}} & \shortstack[r]{C-Eval\\\cite{ceval}} & \shortstack[r]{GPQA\\\cite{gpqa}} & \shortstack[r]{Super\\\cite{supergpqa}} \\
\midrule
\textcolor{YANblue}{\textbf{EdgeRock-4B}} & \textbf{88.2} & \textbf{79.8} & \textbf{92.0} & 73.3 & 74.6 & 64.6 & \textbf{60.0} \\
Qwen3.5-4B~\cite{qwen35} & 85.2 & 77.7 & 89.8 & \textbf{83.9} & \textbf{84.0} & \textbf{77.3} & 52.8 \\
Gemma4-E2B-it~\cite{gemma4} & 72.2 & 47.0 & 78.2 & 60.2 & 61.2 & 46.5 & 34.1 \\
MiniCPM5-2B~\cite{minicpm5} & 79.6 & 65.3 & 85.3 & 79.2 & 78.3 & 67.7 & 40.9 \\
LFM2.5-2.6B~\cite{lfm25config} & 75.3 & 59.9 & 81.7 & 57.2 & 59.5 & 58.2 & 27.1 \\
Qwen3.5-2B~\cite{qwen35} & 72.5 & 62.8 & 79.5 & 68.2 & 68.7 & 54.5 & 39.3 \\
Granite4.2-3B~\cite{granite42} & 77.2 & 66.4 & 81.6 & 59.1 & 60.8 & 58.6 & 39.5 \\
Phi-4-mini-reasoning~\cite{phi4mini} & 77.5 & 38.5 & 78.3 & 45.3 & 42.4 & 48.0 & 32.2 \\
Gemma4-E4B-it~\cite{gemma4} & 81.2 & 69.0 & 86.4 & 68.7 & 68.6 & 55.6 & 41.1 \\
Nemotron-3-Nano-4B~\cite{nemotron3nano} & 75.8 & 56.8 & 81.5 & 54.6 & 56.4 & 43.4 & 37.4 \\
Granite4.2-8B~\cite{granite42} & 78.6 & 72.0 & 86.3 & 55.1 & 58.4 & 61.6 & 45.7 \\
MiniCPM4.1-8B~\cite{minicpm41} & 82.7 & 67.0 & 87.8 & 80.2 & 79.0 & 51.5 & 40.6 \\
\bottomrule
\end{tabularx}
\end{table}
}

\newcommand{\BoundedReasoningResults}{%
\begin{table}[!htbp]
\centering
\caption{Bounded-state commonsense and reasoning scores. Wino. denotes WinoGrande; Hella. denotes HellaSwag.}
\label{tab:reasoning}
\fontsize{9.3}{11.5}\selectfont
\setlength{\tabcolsep}{3pt}
\renewcommand{\arraystretch}{1.12}
\begin{tabularx}{\textwidth}{@{}Y r r r r r r @{}}
\toprule
Model & \shortstack[r]{Wino.\\\cite{winogrande}} & \shortstack[r]{Hella.\\\cite{hellaswag}} & \shortstack[r]{ARC-C\\\cite{arc}} & \shortstack[r]{ARC-E\\\cite{arc}} & \shortstack[r]{BBH\\\cite{bbh}} & \shortstack[r]{BBEH\\\cite{bbeh}} \\
\midrule
\textcolor{YANblue}{\textbf{EdgeRock-4B}} & 77.9 & 64.9 & \textbf{95.5} & \textbf{98.0} & \textbf{85.4} & \textbf{28.9} \\
ARWKV-R1 1.5B~\cite{arwkv} & 31.4 & 7.4 & 19.6 & 22.3 & 9.8 & 1.6 \\
RWKV-7 G1i 1.5B~\cite{rwkvg1} & 46.6 & 33.4 & 68.6 & 83.0 & 27.8 & 4.6 \\
RWKV-7 G1j 1.5B~\cite{rwkvg1} & 49.3 & 24.6 & 66.2 & 80.8 & 31.4 & 5.0 \\
RWKV-7 G1i 2.9B~\cite{rwkvg1} & 57.4 & 62.7 & 81.9 & 92.2 & 30.3 & 4.5 \\
RWKV-7 G1j 2.9B~\cite{rwkvg1} & 58.1 & 65.9 & 83.3 & 91.6 & 39.5 & 6.5 \\
AHN-GDN Qwen2.5 3B~\cite{ahn} & 61.7 & 52.3 & 76.6 & 86.1 & 39.5 & 4.6 \\
ARWKV-R1 7B~\cite{arwkv} & 34.4 & 10.0 & 47.4 & 56.8 & 16.6 & 2.4 \\
Falcon3-Mamba 7B Instruct~\cite{falcon3} & 57.5 & 38.6 & 74.8 & 87.6 & 32.8 & 3.4 \\
QRWKV7 7B Instruct~\cite{radlads} & 61.8 & 48.0 & 86.3 & 95.4 & 37.4 & 1.9 \\
xLSTM 7B~\cite{xlstm7b} & 28.3 & 7.9 & 25.1 & 28.2 & 12.8 & 3.3 \\
RWKV-7 G1i 7.2B~\cite{rwkvg1} & 60.4 & 70.2 & 84.5 & 94.5 & 38.8 & 7.0 \\
RWKV-7 G1j 7.2B~\cite{rwkvg1} & 59.4 & 71.7 & 83.7 & 94.5 & 41.4 & 7.7 \\
RecurrentGemma 9B IT~\cite{recurrentgemma} & 56.4 & 45.8 & 73.3 & 87.2 & 35.5 & 3.2 \\
RWKV-7 G1j 13.3B~\cite{rwkvg1} & 72.1 & \textbf{77.4} & 91.8 & 96.3 & 48.7 & 7.7 \\
AHN-GDN Qwen2.5 14B~\cite{ahn} & 74.0 & 67.9 & 91.5 & 94.0 & 52.0 & 5.3 \\
Phi-3 Medium 4K Instruct~\cite{phi3window} & \textbf{79.6} & 74.0 & 89.8 & 95.3 & 53.3 & 3.1 \\
\bottomrule
\end{tabularx}
\end{table}
}

\newcommand{\GrowingReasoningResults}{%
\begin{table}[!htbp]
\centering
\caption{Commonsense and reasoning comparison with growing-history models. Wino. denotes WinoGrande; Hella. denotes HellaSwag.}
\label{tab:growing-reasoning}
\fontsize{9.3}{11.5}\selectfont
\setlength{\tabcolsep}{3pt}
\renewcommand{\arraystretch}{1.12}
\begin{tabularx}{\textwidth}{@{}Y r r r r r r @{}}
\toprule
Model & \shortstack[r]{Wino.\\\cite{winogrande}} & \shortstack[r]{Hella.\\\cite{hellaswag}} & \shortstack[r]{ARC-C\\\cite{arc}} & \shortstack[r]{ARC-E\\\cite{arc}} & \shortstack[r]{BBH\\\cite{bbh}} & \shortstack[r]{BBEH\\\cite{bbeh}} \\
\midrule
\textcolor{YANblue}{\textbf{EdgeRock-4B}} & 77.9 & 64.9 & 95.5 & 98.0 & 85.4 & 28.9 \\
Qwen3.5-4B~\cite{qwen35} & \textbf{87.1} & \textbf{80.6} & \textbf{96.8} & \textbf{98.4} & 86.2 & \textbf{34.6} \\
Gemma4-E2B-it~\cite{gemma4} & 63.2 & 50.9 & 89.2 & 94.8 & 80.3 & 11.6 \\
MiniCPM5-2B~\cite{minicpm5} & 69.8 & 52.2 & 91.9 & 97.0 & 82.1 & 25.4 \\
LFM2.5-2.6B~\cite{lfm25config} & 67.9 & 40.7 & 91.6 & 96.5 & 76.7 & 13.1 \\
Qwen3.5-2B~\cite{qwen35} & 71.5 & 59.3 & 91.4 & 95.5 & 79.1 & 14.9 \\
Granite4.2-3B~\cite{granite42} & 74.0 & 67.4 & 92.4 & 97.1 & 85.5 & 24.5 \\
Phi-4-mini-reasoning~\cite{phi4mini} & 75.8 & 62.6 & 90.2 & 94.4 & 72.5 & 12.9 \\
Gemma4-E4B-it~\cite{gemma4} & 76.5 & 53.5 & 94.3 & 97.7 & 84.1 & 17.4 \\
Nemotron-3-Nano-4B~\cite{nemotron3nano} & 65.6 & 46.3 & 91.9 & 96.7 & 79.3 & 10.8 \\
Granite4.2-8B~\cite{granite42} & 76.2 & 63.1 & 90.9 & 94.6 & \textbf{88.9} & 31.5 \\
MiniCPM4.1-8B~\cite{minicpm41} & 78.2 & 49.8 & 93.4 & 97.4 & 86.6 & 14.0 \\
\bottomrule
\end{tabularx}
\end{table}
}

\newcommand{\BoundedCodeInstructionResults}{%
\begin{table}[!htbp]
\centering
\caption{Bounded-state coding and instruction following. LCB denotes LiveCodeBench.}
\label{tab:code-instruction}
\fontsize{9.3}{11.5}\selectfont
\setlength{\tabcolsep}{3pt}
\renewcommand{\arraystretch}{1.12}
\begin{tabularx}{\textwidth}{@{}Y r r r r r @{}}
\toprule
Model & \shortstack[r]{HumanEval\\\cite{humaneval}} & \shortstack[r]{MBPP\\\cite{mbpp}} & \shortstack[r]{LCB v6\\\cite{livecodebench}} & \shortstack[r]{IFEval\\\cite{ifeval}} & \shortstack[r]{IFBench\\\cite{ifbench}} \\
\midrule
\textcolor{YANblue}{\textbf{EdgeRock-4B}} & \textbf{96.7} & \textbf{87.8} & \textbf{61.6} & \textbf{88.0} & \textbf{65.3} \\
ARWKV-R1 1.5B~\cite{arwkv} & 4.3 & 5.4 & 0.3 & 11.1 & 8.3 \\
RWKV-7 G1i 1.5B~\cite{rwkvg1} & 42.1 & 36.6 & 7.2 & 37.0 & 14.7 \\
RWKV-7 G1j 1.5B~\cite{rwkvg1} & 41.5 & 38.6 & 6.3 & 46.2 & 16.0 \\
RWKV-7 G1i 2.9B~\cite{rwkvg1} & 58.5 & 49.2 & 11.6 & 43.1 & 17.0 \\
RWKV-7 G1j 2.9B~\cite{rwkvg1} & 58.5 & 48.2 & 12.1 & 56.4 & 17.3 \\
AHN-GDN Qwen2.5 3B~\cite{ahn} & 67.7 & 49.0 & 14.6 & 58.6 & 26.3 \\
ARWKV-R1 7B~\cite{arwkv} & 16.5 & 21.2 & 2.6 & 10.4 & 8.3 \\
Falcon3-Mamba 7B Instruct~\cite{falcon3} & 34.1 & 39.2 & 9.3 & 64.7 & 18.7 \\
QRWKV7 7B Instruct~\cite{radlads} & 62.2 & 30.2 & 12.4 & 59.0 & 21.3 \\
xLSTM 7B~\cite{xlstm7b} & 36.6 & 9.2 & 2.2 & 34.0 & 14.7 \\
RWKV-7 G1i 7.2B~\cite{rwkvg1} & 65.2 & 59.0 & 18.8 & 53.4 & 17.7 \\
RWKV-7 G1j 7.2B~\cite{rwkvg1} & 72.6 & 58.0 & 21.3 & 66.5 & 18.3 \\
RecurrentGemma 9B IT~\cite{recurrentgemma} & 37.2 & 19.0 & 7.9 & 39.7 & 12.7 \\
RWKV-7 G1j 13.3B~\cite{rwkvg1} & 86.0 & 67.4 & 30.7 & 77.8 & 27.3 \\
AHN-GDN Qwen2.5 14B~\cite{ahn} & 75.6 & 68.6 & 32.9 & 77.4 & 33.3 \\
Phi-3 Medium 4K Instruct~\cite{phi3window} & 73.8 & 59.4 & 19.1 & 61.4 & 22.0 \\
\bottomrule
\end{tabularx}
\end{table}
}

\newcommand{\GrowingCodeInstructionResults}{%
\begin{table}[!htbp]
\centering
\caption{Coding and instruction following comparison with growing-history models. LCB denotes LiveCodeBench.}
\label{tab:growing-code-instruction}
\fontsize{9.3}{11.5}\selectfont
\setlength{\tabcolsep}{3pt}
\renewcommand{\arraystretch}{1.12}
\begin{tabularx}{\textwidth}{@{}Y r r r r r @{}}
\toprule
Model & \shortstack[r]{HumanEval\\\cite{humaneval}} & \shortstack[r]{MBPP\\\cite{mbpp}} & \shortstack[r]{LCB v6\\\cite{livecodebench}} & \shortstack[r]{IFEval\\\cite{ifeval}} & \shortstack[r]{IFBench\\\cite{ifbench}} \\
\midrule
\textcolor{YANblue}{\textbf{EdgeRock-4B}} & 96.7 & 87.8 & 61.6 & 88.0 & 65.3 \\
Qwen3.5-4B~\cite{qwen35} & 86.0 & 71.0 & 75.7 & 89.6 & 60.3 \\
Gemma4-E2B-it~\cite{gemma4} & 90.9 & 79.6 & 59.6 & 82.4 & 29.0 \\
MiniCPM5-2B~\cite{minicpm5} & 97.6 & 94.8 & 83.3 & 86.3 & 64.0 \\
LFM2.5-2.6B~\cite{lfm25config} & 92.9 & 73.6 & 59.7 & 93.3 & 56.2 \\
Qwen3.5-2B~\cite{qwen35} & 54.9 & 35.4 & 25.0 & 81.5 & 43.7 \\
Granite4.2-3B~\cite{granite42} & 97.0 & 92.6 & 75.9 & 92.2 & 74.3 \\
Phi-4-mini-reasoning~\cite{phi4mini} & 79.9 & 64.0 & 35.3 & 45.7 & 14.3 \\
Gemma4-E4B-it~\cite{gemma4} & 93.9 & 79.2 & 69.1 & 84.3 & 34.7 \\
Nemotron-3-Nano-4B~\cite{nemotron3nano} & 90.9 & 72.2 & 66.4 & 89.5 & 57.7 \\
Granite4.2-8B~\cite{granite42} & 96.3 & \textbf{95.4} & \textbf{85.6} & \textbf{94.6} & \textbf{77.0} \\
MiniCPM4.1-8B~\cite{minicpm41} & \textbf{98.2} & 92.2 & 70.4 & 73.4 & 23.7 \\
\bottomrule
\end{tabularx}
\end{table}
}

\newcommand{\BoundedAdditionalResults}{%
\begin{table}[!htbp]
\centering
\caption{Bounded-state reading and multi-step reasoning. DROP: F1; MuSR: task-macro accuracy; LogiQA2: accuracy.}
\label{tab:additional}
\fontsize{9.3}{11.5}\selectfont
\setlength{\tabcolsep}{3pt}
\renewcommand{\arraystretch}{1.12}
\begin{tabularx}{\textwidth}{@{}Y r r r @{}}
\toprule
Model & \shortstack[r]{DROP F1\\\cite{drop}} & \shortstack[r]{MuSR\\\cite{musr}} & \shortstack[r]{LogiQA2\\\cite{logiqa2}} \\
\midrule
\textcolor{YANblue}{\textbf{EdgeRock-4B}} & \textbf{77.2} & \textbf{60.4} & 72.2 \\
ARWKV-R1 1.5B~\cite{arwkv} & 5.6 & 8.6 & 17.5 \\
RWKV-7 G1i 1.5B~\cite{rwkvg1} & 23.0 & 19.3 & 38.7 \\
RWKV-7 G1j 1.5B~\cite{rwkvg1} & 24.9 & 29.1 & 36.1 \\
RWKV-7 G1i 2.9B~\cite{rwkvg1} & 24.6 & 31.9 & 47.6 \\
RWKV-7 G1j 2.9B~\cite{rwkvg1} & 30.7 & 34.5 & 45.4 \\
AHN-GDN Qwen2.5 3B~\cite{ahn} & 32.8 & 37.8 & 50.1 \\
ARWKV-R1 7B~\cite{arwkv} & 11.1 & 15.4 & 26.8 \\
Falcon3-Mamba 7B Instruct~\cite{falcon3} & 19.9 & 29.1 & 41.4 \\
QRWKV7 7B Instruct~\cite{radlads} & 21.2 & 2.6 & 33.2 \\
xLSTM 7B~\cite{xlstm7b} & 29.0 & 19.3 & 20.0 \\
RWKV-7 G1i 7.2B~\cite{rwkvg1} & 33.3 & 49.6 & 53.1 \\
RWKV-7 G1j 7.2B~\cite{rwkvg1} & 34.2 & 48.6 & 56.4 \\
RecurrentGemma 9B IT~\cite{recurrentgemma} & 37.1 & 35.2 & 40.2 \\
RWKV-7 G1j 13.3B~\cite{rwkvg1} & 51.5 & 51.4 & 66.6 \\
AHN-GDN Qwen2.5 14B~\cite{ahn} & 66.5 & 56.4 & \textbf{72.3} \\
Phi-3 Medium 4K Instruct~\cite{phi3window} & 51.5 & 51.1 & 62.7 \\
\bottomrule
\end{tabularx}
\end{table}
}

\newcommand{\GrowingAdditionalResults}{%
\begin{table}[!htbp]
\centering
\caption{Reading and multi-step reasoning comparison with growing-history models. DROP: F1; MuSR: task-macro accuracy; LogiQA2: accuracy.}
\label{tab:growing-additional}
\fontsize{9.3}{11.5}\selectfont
\setlength{\tabcolsep}{3pt}
\renewcommand{\arraystretch}{1.12}
\begin{tabularx}{\textwidth}{@{}Y r r r @{}}
\toprule
Model & \shortstack[r]{DROP F1\\\cite{drop}} & \shortstack[r]{MuSR\\\cite{musr}} & \shortstack[r]{LogiQA2\\\cite{logiqa2}} \\
\midrule
\textcolor{YANblue}{\textbf{EdgeRock-4B}} & 77.2 & 60.4 & 72.2 \\
Qwen3.5-4B~\cite{qwen35} & 76.0 & 64.9 & \textbf{82.2} \\
Gemma4-E2B-it~\cite{gemma4} & 72.9 & 58.9 & 59.4 \\
MiniCPM5-2B~\cite{minicpm5} & 76.0 & \textbf{65.3} & 73.7 \\
LFM2.5-2.6B~\cite{lfm25config} & 72.3 & 60.4 & 63.2 \\
Qwen3.5-2B~\cite{qwen35} & 73.7 & 61.3 & 66.3 \\
Granite4.2-3B~\cite{granite42} & \textbf{83.7} & 65.1 & 69.5 \\
Phi-4-mini-reasoning~\cite{phi4mini} & 50.6 & 56.7 & 58.5 \\
Gemma4-E4B-it~\cite{gemma4} & 72.9 & 62.2 & 72.6 \\
Nemotron-3-Nano-4B~\cite{nemotron3nano} & 72.6 & 58.2 & 67.9 \\
Granite4.2-8B~\cite{granite42} & 76.9 & 62.4 & 73.8 \\
MiniCPM4.1-8B~\cite{minicpm41} & 78.3 & 64.5 & 75.7 \\
\bottomrule
\end{tabularx}
\end{table}
}

\newcommand{\PeerCapabilityResults}{%
\begin{table}[!htbp]
\centering\small
\caption{Compact-model capability (\%). EdgeRock keeps constant persistent history; models below the divider are growing-history models. AIME: 2024--2026 mean. Bold: column maximum.}
\label{tab:peer-capability}
\setlength{\tabcolsep}{4pt}
\renewcommand{\arraystretch}{1.18}
\begin{tabular*}{\linewidth}{@{\extracolsep{\fill}}lrrrrrr@{}}
\toprule
Model & AIME & \shortstack[r]{MMLU\\Pro} & \shortstack[r]{Super\\GPQA} & \shortstack[r]{GPQA\\Diamond} & \shortstack[r]{LCB\\v6} & IFEval\\
\midrule
\textcolor{YANblue}{\textbf{EdgeRock-4B}} & 82.9 & \textbf{79.8} & \textbf{60.0} & 64.6 & 61.6 & 88.0 \\
\midrule
Qwen3.5-4B~\cite{qwen35} & 86.1 & 77.7 & 52.8 & \textbf{77.3} & 75.7 & 89.6 \\
Gemma4-E4B-it~\cite{gemma4} & 40.1 & 69.0 & 41.1 & 55.6 & 69.1 & 84.3 \\
Nemotron-3-Nano-4B~\cite{nemotron3nano} & 57.1 & 56.8 & 37.4 & 43.4 & 66.4 & 89.5 \\
Phi-4-mini-reasoning~\cite{phi4mini} & 38.9 & 38.5 & 32.2 & 48.0 & 35.3 & 45.7 \\
MiniCPM5-2B~\cite{minicpm5} & \textbf{87.8} & 65.3 & 40.9 & 67.7 & \textbf{83.3} & 86.3 \\
LFM2.5-2.6B~\cite{lfm25config} & 56.2 & 59.9 & 27.1 & 58.2 & 59.7 & \textbf{93.3} \\
\bottomrule
\end{tabular*}
\end{table}
}

\newcommand{\EdgePeakBatchRatio}{6.42}
\newcommand{\EdgeBatchRatio}{4.94}
\newcommand{\EdgeLFMRatio}{7.18}
\newcommand{\EdgeLongTPS}{233.7}
\newcommand{\EdgeLongMemory}{16.02}
\newcommand{\EdgeSingleRatio}{1.54}
\newcommand{\EdgeWeightedTPS}{12,497}
\newcommand{\EdgePerSequenceTPS}{24.4}
\newcommand{\EdgeBatchMemory}{61.26}
\newcommand{\EdgeShortTPS}{229.6}
\newcommand{\EdgeShortMemory}{14.27}
\newcommand{\EdgeShortBatchTPS}{12,481}
\newcommand{\PeerThroughputResults}{%
\begin{table}[!htbp]
\centering\small
\caption{Output-token throughput with 32K output on one Hopper 80GB. Only output tokens are counted; elapsed time includes prefill. Columns give input length; cells show output tokens/s (best batch size). Bold marks the highest throughput in each column.}
\label{tab:peer-throughput}
\setlength{\tabcolsep}{4pt}
\begin{tabularx}{\linewidth}{@{}>{\raggedright\arraybackslash}p{.22\linewidth}*{4}{>{\raggedleft\arraybackslash}X}@{}}
\toprule
Model & 4K & 8K & 16K & 32K\\
\midrule
EdgeRock-4B & \textbf{12,479} (512) & \textbf{12,325} (512) & \textbf{12,330} (512) & \textbf{12,124} (512)\\
Qwen3.5-4B & 3,103 (64) & 2,853 (64) & 2,472 (64) & 1,889 (64)\\
Gemma 4 E4B & 3,940 (128) & 3,432 (128) & 2,880 (128) & 2,196 (64)\\
MiniCPM5-2B & 3,152 (64) & 2,938 (64) & 2,575 (32) & 2,117 (32)\\
LFM2.5-2.6B & 1,673 (64) & 1,771 (64) & 1,695 (32) & 1,727 (64)\\
\bottomrule\end{tabularx}\end{table}}
\newcommand{\BatchScalingResults}{%
\begin{table}[!htbp]
\centering\small
\caption{Weighted decoding and peak GPU memory across 16 input/output pairs (4K--32K) on one Hopper 80GB.}
\label{tab:batch-scaling}
\begin{tabularx}{\linewidth}{@{}Yrrrr@{}}
\toprule
Batch size & Decode, tokens/s & Per sequence, tokens/s & Allocated, GiB & Reserved, GiB\\
\midrule
32 & 4,128 & 129.0 & 18.94 & 19.49\\
64 & 6,685 & 104.5 & 20.16 & 21.80\\
128 & 9,049 & 70.7 & 26.74 & 27.92\\
256 & 11,231 & 43.9 & 39.90 & 41.77\\
320 & 11,074 & 34.6 & 46.49 & 50.69\\
384 & 12,099 & 31.5 & 53.06 & 57.27\\
512 & 12,497 & 24.4 & 61.26 & 66.97\\
\bottomrule\end{tabularx}\end{table}}
\newcommand{\ShortScalingResults}{%
\begin{table}[!htbp]
\centering\small
\caption{Decode throughput and peak GPU allocation at 4K input and 2K output on one Hopper 80GB.}
\label{tab:short-scaling}
\begin{tabularx}{\linewidth}{@{}Yrrr@{}}
\toprule
Batch size & Decode, tokens/s & Per sequence, tokens/s & Allocated, GiB\\
\midrule
1 & 230 & 229.6 & 14.27\\
2 & 358 & 178.8 & 14.27\\
4 & 692 & 172.9 & 14.27\\
8 & 1,322 & 165.2 & 14.40\\
16 & 2,442 & 152.6 & 15.21\\
32 & 4,053 & 126.7 & 16.84\\
64 & 6,698 & 104.7 & 20.10\\
128 & 9,066 & 70.8 & 26.60\\
256 & 11,100 & 43.4 & 39.62\\
320 & 11,062 & 34.6 & 46.14\\
384 & 12,086 & 31.5 & 52.64\\
512 & 12,481 & 24.4 & 60.71\\
\bottomrule\end{tabularx}\end{table}}

\begin{document}
\fancypagestyle{brandtitle}{%
\fancyhf{}
\fancyhead[L]{\raisebox{-7pt}{\includegraphics[height=21pt]{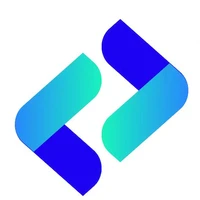}}\hspace{5pt}\raisebox{-4pt}{{\sffamily\bfseries\fontsize{9}{10}\selectfont\shortstack[l]{ROCK\\[-1pt]MATRIX}}}}
\fancyfoot[C]{\thepage}
\renewcommand{\headrulewidth}{0.3pt}
\renewcommand{\headruleskip}{2pt}
}
\thispagestyle{brandtitle}
\begin{center}
{\LARGE\bfseries EdgeRock: Strong Reasoning at Constant Memory\par}
\vspace{12pt}
\begingroup
\hypersetup{urlcolor=black}
\renewcommand{\arraystretch}{1.0}
\begin{tabular*}{0.84\linewidth}{@{\extracolsep{\fill}}ccc@{}}
Huishan Ji & Weiming Zhang & Qirui Ye \\
{\small\href{mailto:jihs@stonehill-tech.com}{jihs@stonehill-tech.com}} &
{\small\href{mailto:zhangwm@stonehill-tech.com}{zhangwm@stonehill-tech.com}} &
{\small\href{mailto:qiruiy@andrew.cmu.edu}{qiruiy@andrew.cmu.edu}} \\[5pt]
Jinchao Xu & Yike Guo & Hua Xu\textsuperscript{*} \\
{\small\href{mailto:jinchao.xu@kaust.edu.sa}{jinchao.xu@kaust.edu.sa}} &
{\small\href{mailto:yikeguo@ust.hk}{yikeguo@ust.hk}} &
{\small\href{mailto:huaxu@ust.hk}{huaxu@ust.hk}}
\end{tabular*}
\endgroup
\par\vspace{8pt}
{\sffamily\footnotesize
\href{https://github.com/rockmatrix-ai/EdgeRock}{\raisebox{-2.2pt}{\includegraphics[height=10pt]{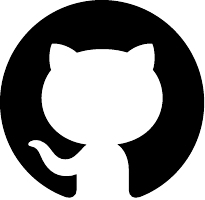}}\hspace{5pt}github.com/rockmatrix-ai/EdgeRock}\hspace{2.2em}
\href{https://huggingface.co/rockmatrix-ai/EdgeRock}{\raisebox{-2.2pt}{\includegraphics[height=10pt]{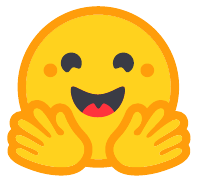}}\hspace{5pt}huggingface.co/rockmatrix-ai/EdgeRock}\par}
\end{center}
\begingroup
\renewcommand{\thefootnote}{\fnsymbol{footnote}}
\footnotetext[1]{Corresponding author.}
\endgroup
\begin{abstract}
We present EdgeRock, a family of multimodal models that deliver strong reasoning at constant memory. Most models keep reasoning history in a \mbox{key--value} (KV) cache that grows with every token, consuming accelerator memory, limiting concurrent requests, and competing with perception and control on edge devices. With fixed history, memory is allocated once and never runs out, however long the model reasons. Constant memory alone is not new: bounded-state models (recurrent, state-space, and sliding-window) already achieve it, but often at the cost of reasoning, because they cannot keep the distant details that later reasoning needs. We therefore introduce \emph{Rock Memory Anchors} (RMA): the model learns what is worth remembering and stores it as KV records in a fixed-size bank, directly readable however far back the source lies. EdgeRock-4B, our first open-source release, supports text and image reasoning and reaches 82.9\% mean pass@1 on AIME 2024--2026, while the bounded-state models we evaluate reach at most 18.8\%, even at 14B scale. This result is on par with leading compact models whose memory grows with reasoning. Compared with its source model, EdgeRock-4B uses up to 99.1\% less KV cache and delivers up to \EdgePeakBatchRatio{}$\times$ higher end-to-end batch throughput.
\end{abstract}
\begin{figure}[!ht]
\centering
\includegraphics[width=0.96\linewidth]{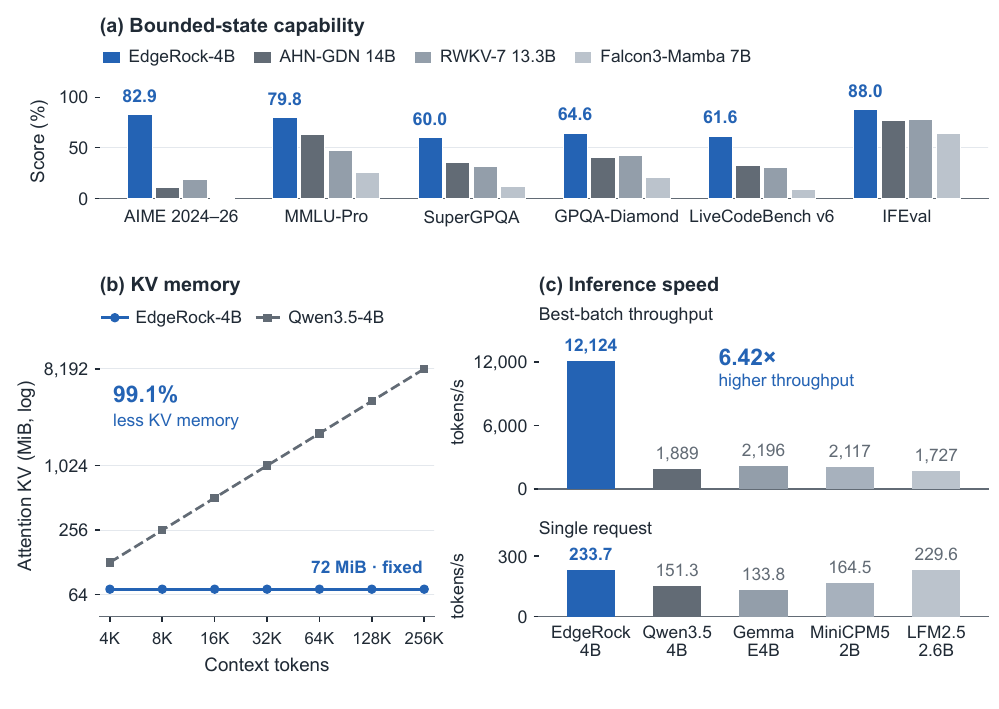}
\caption{\textbf{Reasoning at constant memory.} \textbf{(a)} EdgeRock-4B leads the displayed 7B--14B bounded-state models on six text metrics; AIME averages 2024--2026 equally. \textbf{(b)} Configuration-derived BF16 attention KV capacity per sequence (log axes). EdgeRock uses 72\,MiB, 99.1\% less than Qwen3.5-4B at 256K. \textbf{(c)} Top: output-token throughput for 32K input and 32K output with repeated prompts, including prefill time, at each model's best measured batch (Appendix Table~\ref{tab:peer-throughput}); EdgeRock achieves \EdgePeakBatchRatio{}$\times$ Qwen3.5-4B's throughput. Bottom: cumulative decode rate for 128K input and 128K output at batch size 1. All runs use BF16 weights and KV on one Hopper 80GB GPU.}
\label{fig:overview}
\end{figure}
\clearpage

\section{Introduction}
\label{sec:thesis}

Long reasoning turns memory into the deployment bottleneck. Under full attention~\cite{transformer}, every generated token adds to the key--value (KV) cache, so a model that reasons for 100K tokens must also store 100K tokens of history. We call such models, whose memory grows with reasoning, \emph{growing-history} models. In data centers, their caches consume limited high-bandwidth memory (HBM), cutting the number of requests each accelerator can serve and raising serving costs~\cite{pagedattention}.

For many deployments, a smaller cache is not enough: the cache must be fixed. On edge devices, robots and vehicles must reserve limited memory for perception, planning, and control~\cite{edge_llm}; dedicated chips go further and fix on-chip capacity at design time. A growing cache forces a choice between reserving space for the longest possible trace and cutting reasoning short when memory runs out. A fixed history budget makes that allocation predictable: the same memory serves a sequence at 4K or 256K tokens. What these deployments need is determinism, not only efficiency.

Recent architectures shrink this cache aggressively: multi-head latent attention (MLA) compresses each cached token~\cite{deepseekv2}; hybrids such as Qwen3.5 and Kimi Linear build most layers from Gated DeltaNet (GDN) or Kimi Delta Attention (KDA) and keep full attention in only one layer in four~\cite{qwen35,kimilinear}; DeepSeek Sparse Attention (DSA) reads only a subset of entries~\cite{deepseekv32}, and the compressed sparse attention (CSA) of DeepSeek-V4 stores fewer~\cite{deepseekv4}; and KV quantization methods such as KIVI use fewer bits per entry~\cite{kivi}. Each lowers the cost of a cached token, but the cache still grows with every token the model reads or writes. Shrinking the growth rate moves the memory wall further out; it does not remove it.

\emph{Bounded-state} models already provide this determinism in history storage. Recurrent and state-space models compress history into fixed-size states~\cite{mamba,rwkv7,xlstm}, and sliding-window attention, alone or in recurrent hybrids, keeps only recent tokens~\cite{longformer,recurrentgemma,samba}. None that we evaluate delivers usable reasoning: across 16 bounded-state models up to 14B parameters, none exceeds 18.8\% on AIME 2024--2026 (Section~\ref{sec:capability}). Inference-time eviction can also cap the cache of a growing-history model by keeping attention sinks, recent tokens, or tokens that drew high attention~\cite{streamingllm,h2o,snapkv}, but the model was never trained to reason with only what remains.

Long reasoning keeps returning to details stated thousands of tokens earlier: a definition, a constraint in the problem statement, an intermediate result. A sliding window forgets such a detail once it scrolls out; a recurrent state keeps only a compressed summary in which individual details blur together~\cite{repeatafterme}. The question becomes: \emph{can a model keep the distant details that later reasoning needs within a fixed memory, so that constant memory no longer costs reasoning ability?}

EdgeRock addresses this with \emph{Rock Memory Anchors} (RMA). Like a rock anchor that pins a structure to bedrock, RMA pins critical details in place. Each \emph{anchor} is a learned key--value record held in a fixed-size \emph{anchor bank}. Recurrent state summarizes the evolving context, local attention covers the most recent 1,024 tokens, and RMA keeps the most valuable details from earlier context. A learned writer produces a candidate record for each token, and a learned scorer rates it. Once 256 candidates have accumulated, each KV group compares them with the records already in its bank, keeps the highest-scoring 1,024 records, and discards the rest. Kept records stay separate and unchanged, so any later token can attend to them directly, long after their source has left the window. Local attention and RMA together hold attention KV at 72\,MiB per sequence, however long the model reasons; with recurrent state, everything a sequence must store to continue, its \emph{persistent history}, stays at 97.76\,MiB in the batch runtime.

EdgeRock-4B, our first open release, shows that constant memory and strong reasoning are compatible (Table~\ref{tab:contributions}).

\begin{table}[!htbp]
\centering\small
\caption{EdgeRock-4B at a glance.}
\label{tab:contributions}
\begin{tabularx}{\linewidth}{@{}>{\raggedright\arraybackslash}p{.16\linewidth}YY@{}}
\toprule
Claim & How EdgeRock achieves it & Supporting result\\
\midrule
Constant memory & Recurrent state, a 1,024-token window, and RMA's fixed-size anchor banks. & 72\,MiB of attention KV per sequence at any length; 99.1\% below Qwen3.5-4B at 256K context (Section~\ref{sec:complexity}).\\
\addlinespace[3pt]
Usable reasoning & Learned writing, admission, and retrieval, trained in stages. & 82.9\% on AIME 2024--2026, versus at most 18.8\% for 16 bounded-state models; within the 82.2--87.8\% range of leading compact models (Section~\ref{sec:capability}).\\
\addlinespace[3pt]
RMA is essential & Anchors stay individually retrievable after leaving the window. & Disabling RMA at inference drops long-context query accuracy from 94.1\% to 0.4\% (Section~\ref{sec:diagnostic}).\\
\addlinespace[3pt]
Faster serving & Fixed history per sequence: per-token work does not grow with context, and one GPU runs more sequences at once. & 512 concurrent long-generation sequences on one Hopper 80GB GPU; \EdgeBatchRatio{}$\times$ Qwen3.5-4B's throughput on average, up to \EdgePeakBatchRatio{}$\times$ (Section~\ref{sec:throughput}).\\
\bottomrule
\end{tabularx}
\end{table}

\FloatBarrier

\section{How is EdgeRock built?}
\label{sec:architecture}

\subsection{Decoder layout}

\begin{figure}[!tbp]
\centering
\includegraphics[width=\linewidth]{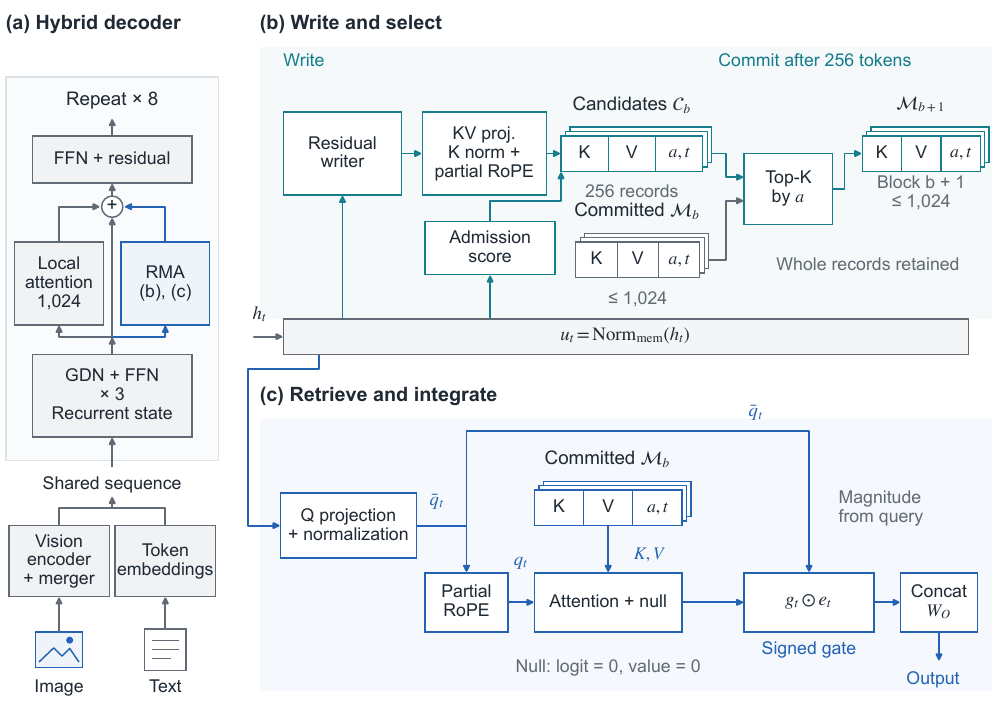}
\caption{\textbf{EdgeRock architecture.} \textbf{(a)} Text and image features share a decoder combining recurrence, local attention, and RMA. \textbf{(b)} Per KV group, admission selects up to 1,024 records from the committed bank and 256 writer-generated candidates. Selections become readable in the next block. \textbf{(c)} Current queries read committed anchors; null attention and a query-conditioned signed gate control their contribution to the residual stream alongside local attention.}
\label{fig:architecture}
\end{figure}

EdgeRock-4B adapts the Qwen3.5-4B language backbone~\cite{qwen35}. It retains 24 Gated DeltaNet (GDN) layers and replaces its eight full-attention layers with a 1,024-position sliding window~\cite{longformer} plus an independent RMA module. The resulting pattern is
\begin{equation}
\bigl[\mathrm{GDN},\ \mathrm{GDN},\ \mathrm{GDN},\ \mathrm{SWA}_{1024}\parallel\mathrm{RMA}\bigr]\times8.
\end{equation}
Every sequence mixer is followed by a feed-forward sublayer; each of the eight RMA modules has its own parameters and state. Image features from the visual encoder and merger join text embeddings in a shared decoder input sequence (Figure~\ref{fig:architecture}a).

GDN uses fixed-dimensional recurrent matrices~\cite{gateddeltanet}. The local window and anchor bank retain separate KV records; a source position may appear in both.

\Needspace{12\baselineskip}
\subsection{Fixed capacity, adaptive content}

EdgeRock keeps the size of its persistent history constant and updates only its contents. Let $t=0,1,\ldots$ index causal sequence positions and $b(t)=\lfloor t/C\rfloor$ identify the current candidate block of $C=256$ positions. Immediately before position $t$,
\begin{align}
\mathcal H_t&=(\mathcal R_t,\mathcal L_t,\mathcal M_{b(t)},\mathcal C_t,\pi_t),\label{eq:history-state}\\
(y_t,\mathcal H_{t+1})&=F_\theta(x_t,\mathbf p_t,\mathcal H_t).\label{eq:history-transition}
\end{align}
Here $x_t$ is the input embedding, $y_t$ is the decoder output, and $\mathbf p_t$ contains the rotary coordinates. Across the decoder, $\mathcal R_t$ collects recurrent states, $\mathcal L_t$ local KV, $\mathcal M_{b(t)}$ committed anchor banks, $\mathcal C_t$ pending candidates, and $\pi_t$ progress metadata.

\subsection{Integration into the reasoning computation}

Suppressing layer indices, each local-plus-RMA layer integrates retrieved information into the residual stream alongside local attention:
\begin{align}
z_t &= h_t+\mathrm{SWA}(\mathrm{Norm}_{\mathrm{local}}(h_t))+\mathrm{RMA}(h_t,\mathcal M_{b(t)}),\\
h_t^{\mathrm{out}} &= z_t+\mathrm{MLP}(\mathrm{Norm}_{\mathrm{ffn}}(z_t)).
\end{align}
RMA uses independent normalization, writing, admission, query, key, and value projections, and output projection. The residual $z_t$ carries local and retrieved information to later layers.

\begin{table}[!htbp]
\centering\small
\caption{EdgeRock-4B configuration.}
\label{tab:architecture}
\begin{tabularx}{\linewidth}{@{}YlYl@{}}
\toprule
Quantity & Value & Quantity & Value\\
\midrule
Model scale & 4B & RMA parameters & 372M\\
Recurrent / local-plus-RMA layers & 24 / 8 & Residual width & 2,560\\
Attention query heads / KV groups & 16 / 4 & Attention head dimension & 256\\
Local window $W$ & 1,024 & Committed capacity $K$ & 1,024\\
Candidate block $C$ & 256 & RMA KV precision & BF16\\
\bottomrule
\end{tabularx}
\end{table}

\FloatBarrier
\subsection{Staged conversion}
\label{sec:training}
EdgeRock-4B is converted from Qwen3.5-4B in four stages. RMA adaptation (M0) teaches the new RMA modules to retain and retrieve information needed for later predictions. Joint adaptation (R) trains RMA together with the recurrent and local-attention backbone, so downstream layers learn to use what is retrieved. Supervised post-training (S) develops general-purpose responses with the adapted history mechanism, and reinforcement learning with verifiable rewards (RLVR)~\cite{tulu3,deepseekr1} refines responses using feedback on answer correctness and task completion. Section~\ref{sec:diagnostic} tracks memory accuracy after each stage (Figure~\ref{fig:memory}a).

\section{How does RMA preserve distant information?}
\label{sec:memory-algorithm}

At each token, RMA reads the committed anchor bank and produces a candidate record from the same causal input. At each block boundary, admission selects records from the bank and the new candidates; the updated bank becomes readable in the next block. Local attention covers recent tokens while candidates accumulate. Algorithm~\ref{alg:anchor-update} in Appendix~\ref{sec:streaming-execution} gives the complete per-token procedure.

\subsection{Independent memory writer}
\label{sec:write}

Let $u_t=\mathrm{Norm}_{\mathrm{mem}}(h_t)\in\mathbb R^D$, with $D=2{,}560$. An independent residual writer with a SwiGLU-style gated projection~\cite{glu} constructs a storage representation:
\begin{equation}
w_t = u_t+W_{\mathrm{down}}\!\left[\mathrm{SiLU}(W_{\mathrm{gate}}u_t)\odot W_{\mathrm{up}}u_t\right].
\end{equation}
RMA KV projections use the writer output $w_t$; query projections use $u_t$. The writer has an intermediate width of 2,560~\cite{yanchorcode,yanchorconfig}.

\subsection{Causal admission in four independent KV groups}

A small network assigns a retention priority to each KV group:
\begin{equation}
a_t = W_{a,2}\,\mathrm{SiLU}(W_{a,1}u_t),\qquad a_t\in\mathbb R^4.
\end{equation}
Bias-free projections of size $2560\!\rightarrow\!256\!\rightarrow\!4$ compute FP32 scores from the current causal context~\cite{yanchorcode,yanchorconfig}. Each group independently retains its highest-scoring records; source positions may overlap across groups.

\subsection{Blockwise commits and causal visibility}

Every 256 tokens, the committed bank and new candidates compete for 1,024 slots per KV group.

For block $\mathcal B_b=\{bC,\ldots,(b+1)C-1\}$, group $g$, capacity $K=1024$ and block size $C=256$,
\begin{align}
\mathcal C_{b,g}&=\{(k_{t,g},v_{t,g},a_{t,g},t):t\in\mathcal B_b\},\\
\mathcal M_{b+1,g}&=\operatorname{TopK}_{a_g}(\mathcal M_{b,g}\cup\mathcal C_{b,g};K).
\end{align}
During block $b$, queries read $\mathcal M_{b,g}$; selected candidates become readable in block $b+1$. The 1,024-position local window covers current-block interactions while the 256-token candidate buffer fills. Each update inspects at most $K+C=1280$ records per group and then reuses the candidate buffer.

The operator $\operatorname{TopK}(A;K)$ retains up to $K$ records, ordered by descending score and then earlier causal position. With fixed write-time scores, the committed bank satisfies the exact history invariant
\begin{equation}
\mathcal M_{b,g}=\operatorname{TopK}_{a_g}\!\left(\bigcup_{r=0}^{b-1}\mathcal C_{r,g};K\right),
\qquad \mathcal M_{0,g}=\varnothing.
\label{eq:history-topk}
\end{equation}
Every discarded record already has at least $K$ records ahead of it, so later candidates cannot bring it into the global top $K$. The update therefore maintains the highest-priority records from all completed blocks using bounded storage. Capacity limits record count, not retention age: a record remains directly addressable until displaced by admission.

\Needspace{10\baselineskip}
\subsection{Query-dependent retrieval from the anchor bank}
\label{sec:read}

A later query can address any retained record, including one whose source position has left the local window. Each module uses 16 query heads over four KV groups~\cite{gqa}, with head dimension $d=256$. Queries and keys use learned RMS normalization~\cite{rmsnorm,yanchorcode,yanchorconfig}:
\begin{equation}
\operatorname{RMS}_{\gamma}(x)=\frac{\gamma\odot x}{\sqrt{d^{-1}\sum_{i=1}^{d}x_i^2+10^{-6}}}.
\end{equation}
The query reads the current representation $u_t$; keys and values store the writer representation $w_t$:
\begin{align}
\bar q_{t,h}&=\operatorname{RMS}_{\gamma_Q}(W_{Q,h}u_t),
&q_{t,h}&=R(\mathbf p_t)\bar q_{t,h},\\
k_{t,g}&=R(\mathbf p_t)\operatorname{RMS}_{\gamma_K}(W_{K,g}w_t),
&v_{t,g}&=W_{V,g}w_t.
\label{eq:memory-qkv}
\end{align}
Each projection $W_{Q,h},W_{K,g},W_{V,g}\in\mathbb R^{d\times D}$. Query and key gains are separate and shared across heads or groups within a module, with effective gain $\gamma=1+\omega$. The partial rotary transform~\cite{roformer} $R(\mathbf p_t)$ rotates 64 of 256 coordinates with base $10^7$; values are stored without this transform or Q/K normalization. Text-only input uses $\mathbf p_t=(t,t,t)$, while multimodal input uses the backbone's spatial and temporal coordinates (Appendix~\ref{sec:streaming-execution}).

For head $h$, assigned to KV group $g(h)$, let $\mathcal I_{b,g}$ index the records in $\mathcal M_{b,g}$ and define $\ell_{tj,h}=q_{t,h}^{\top}k_{j,g(h)}/\sqrt d$. With $b=b(t)$, the committed-bank read is
\begin{equation}
e_{t,h}=\sum_{j\in\mathcal I_{b,g(h)}}
\frac{\exp(\ell_{tj,h})}
{1+\sum_{i\in\mathcal I_{b,g(h)}}\exp(\ell_{ti,h})}\,v_{j,g(h)}.
\end{equation}
Each head reads at most 1,024 records. Query heads share each group's KV bank but compute distinct attention weights.

\paragraph{Retrieval content and strength.}
The null item has zero logit and value. Its attention mass is $\alpha_{0,t,h}=(1+Z_{t,h})^{-1}$, with $Z_{t,h}=\sum_{j\in\mathcal I_{b,g(h)}}\exp(\ell_{tj,h})$. For a nonempty bank, the read factors into
\begin{equation}
e_{t,h}=(1-\alpha_{0,t,h})\,\bar v_{t,h},
\qquad
\bar v_{t,h}=\sum_{j\in\mathcal I_{b,g(h)}}
\frac{\exp(\ell_{tj,h})}{Z_{t,h}}\,v_{j,g(h)}.
\label{eq:retrieval-factorization}
\end{equation}
The query determines both the retrieved content $\bar v_{t,h}$ and the total mass assigned to memory. The read lies in the convex hull of the stored values and zero; an empty bank returns exactly zero.

\subsection{Query-conditioned memory output}
The output gate acts on each retrieved head:
\begin{align}
g_{t,h}&=\tanh(s_h)\,2\sigma\!\left(\frac{\bar q_{t,h}^{\top}b_h}{\sqrt d}+c_h\right),\\
\mathrm{RMA}(h_t,\mathcal M_{b(t)})&=W_O\,\operatorname{concat}_{h}\!\left(g_{t,h}e_{t,h}\right).
\end{align}
Here $\bar q$ is the normalized query before rotary encoding; $s_h,c_h\in\mathbb R$, $b_h\in\mathbb R^d$, and $W_O\in\mathbb R^{D\times16d}$ are learned parameters. The gate is a signed multiplier in $(-2,2)$. Its sign comes from the learned scale $\tanh(s_h)$; the query-conditioned sigmoid modulates its magnitude.

Admission determines which records are available, retrieval determines their content and attention mass, and the signed gate controls their contribution to the residual stream.

\paragraph{Relation to other retained-memory designs.}
Several designs also keep a small set of key--value records next to a compressed or local history. AnLLM trains a model to summarize each text segment into an anchor token and keep only those tokens' keys and values~\cite{anllm}; its anchors still accumulate as the sequence grows. HOLA adds to a linear-attention model a bounded cache of recurrent keys and values ranked by a fixed, parameter-free priority~\cite{hola}, and LTE learns per-head scores that decide which tokens to retain as they leave a sliding window~\cite{lte}. RMA differs in three ways: an independent writer produces dedicated records rather than copying existing keys and values; a causal scorer admits them into a fixed-size bank through per-group top-$K$ commits every 256 tokens; and a separately normalized, gated read integrates them alongside local attention in a converted reasoning model.

\section{What stays constant as context grows?}
\label{sec:complexity}

Fixed-size recurrent state, the local window, and anchor banks bound both stored history and per-token history access.

\subsection{Memory that does not grow with sequence length}

For prompt length $P$ and generated length $N$, hold model dimensions, layer counts, precision, local window $W$, memory capacity $K$ and block size $C$ fixed. Storage counts fixed-width tensor elements and metadata words. The complete persistent history allocation is
\begin{equation}
S_{\mathrm{history}}=S_{\mathrm{GDN}}+S_{\mathrm{SWA}}+S_{\mathrm{committed}}+S_{\mathrm{candidates}}+S_{\mathrm{metadata}}.
\end{equation}
Recurrent matrices, the local ring, committed anchor banks, pending candidates, scores and indices all have bounded capacity. The history allocation is consequently $O(1)$ in $P+N$ for each sequence.

\begin{align}
S_{\mathrm{committed}}&=2\times8\times4\times1024\times256\times2\ \mathrm{bytes}=32\ \mathrm{MiB},\\
S_{\mathrm{candidates}}&=2\times8\times4\times256\times256\times2\ \mathrm{bytes}=8\ \mathrm{MiB}.
\end{align}
The factors count keys and values, modules, KV groups, records, head width and two-byte BF16 elements. RMA uses \textbf{40 MiB of KV per sequence}. A window of 1,024 positions adds at most 32\,MiB, giving \textbf{72\,MiB of attention KV per sequence}. With the same eight attention layers, four KV heads, head width, and BF16 precision, a 256K-token full-attention cache requires $32\,\mathrm{KiB}\times262{,}144=8$\,GiB. The attention KV reduction is $1-72/8192\approx\mathbf{99.1\%}$. Table~\ref{tab:state-budget} also accounts for recurrent state and auxiliary history tensors.
\begin{table}[!htbp]
\centering\small
\caption{Persistent history per sequence. Every entry is fixed in size, independent of context length.}
\label{tab:state-budget}
\begin{tabularx}{\linewidth}{@{}Ylrr@{}}
\toprule
Component & Storage type & MiB per sequence & Share of total\\
\midrule
\textbf{Attention KV} &  & \textbf{71.97} & \textbf{73.6\%}\\
\quad Local window (1,023 preceding positions) & BF16 & 31.97 & 32.7\%\\
\quad \textbf{RMA} & BF16 & \textbf{40.00} & \textbf{40.9\%}\\
\qquad Committed anchors &  & 32.00 & 32.7\%\\
\qquad Pending candidates &  & 8.00 & 8.2\%\\
\addlinespace[3pt]
\textbf{Recurrent state} &  & \textbf{25.50} & \textbf{26.1\%}\\
\quad Recurrent matrices & FP16 & 24.00 & 24.6\%\\
\quad Convolution state & BF16 & 1.50 & 1.5\%\\
\addlinespace[3pt]
\textbf{Metadata} (scores, positions, validity, counters) &  & \textbf{0.29} & \textbf{0.3\%}\\
\midrule
\textbf{Total} &  & \textbf{97.76} & \textbf{100.0\%}\\
\bottomrule
\end{tabularx}
\end{table}

\subsection{Work per generated token}
Each token updates fixed recurrent states, reads at most $W$ local positions and $K$ anchors per group, and performs fixed-width projections and feed-forward computation, with fixed non-attention cost $c_{\mathrm{fixed}}$. A block commit selects from at most $K+C$ records. Using sorting as a conservative selection bound gives
\begin{equation}
T_{\mathrm{decode}}(N)=O(Nc_{\mathrm{fixed}})+O\!\left(N(W+K)\right)
+O\!\left(\left\lceil\frac{N}{C}\right\rceil(K+C)\log(K+C)\right)=O(N).
\end{equation}
Streamed, bounded-chunk prefill gives \textbf{$O(P+N)$ total stream-processing work}. Full attention requires $O(PN+N^2)$ attention work during decode and $O(P+N)$ history storage. Latent compression, cross-token compression, sparse reads, and hybrid layer ratios lower the constants in these terms but leave history storage linear in $P+N$; EdgeRock removes that dependence.

\subsection{Inference cache and request concurrency}
History storage grows with the number of concurrent requests $B$, not with their length. The optimized runtime implements these fixed shapes with CUDA Graphs, packed projections, fused updates, and physical sliding-window caches, and keeps each request's history in its own slot. Requests that repeat a prompt can share its prefill, but each continuation keeps its own state.

Prefill chunks and resumed requests carry recurrent state, local KV, committed records, pending candidates, and positions together; Appendix~\ref{sec:streaming-execution} specifies continuation semantics.

\FloatBarrier
\section{Evaluation design}
\label{sec:evaluation}

We compare 28 models on 29 text benchmarks and evaluate visual reasoning, distant memory, and long-sequence inference. Complete text scores appear in Appendices~\ref{sec:full-results} and~\ref{app:growing-results}.

\subsection{Comparison objectives}
The evaluation makes five comparisons, each isolating one question: 16 bounded-state models, including three 14B-class releases, show where the constant-memory class stands; the source model, Qwen3.5-4B, measures what the conversion costs; compact growing-history models, including MiniCPM5-2B and LFM2.5-2.6B, place EdgeRock-4B among models of its size; enabling and disabling RMA, together with paired changes to distant evidence, tests whether distant answers depend on the anchors; and four compact models provide the reference for decode rate, batch throughput, and GPU memory.

\paragraph{Bounded-state comparison class.}
The comparison covers recurrent and state-space models, recurrent/local hybrids~\cite{rwkv7,rwkvg1,radlads,arwkv,ahn,falcon3,recurrentgemma,xlstm7b}, and windowed attention. Phi-3 Medium 4K uses a configured 2,047-token sliding window~\cite{phi3window}.

\subsection{Metrics and execution profiles}
Most tasks report accuracy; DROP reports F1. Scores follow each benchmark's prescribed subject or subtask aggregation, and the overall mean weights all 29 primary metrics equally. AIME averages the 2024--2026 editions equally. In tables that compare three or more models, bold marks the best result for each metric, including ties; paired comparisons with Qwen3.5-4B report retention instead.

Memory and visual evaluations use reference arithmetic; runtime experiments use the optimized profiles in Section~\ref{sec:systems}. Appendix~\ref{app:protocol} specifies sampling, scoring, and source configurations.

\section{Does strong reasoning survive constant memory?}
\label{sec:capability}
EdgeRock-4B brings bounded-state reasoning to a usable level. It ranks first among 17 bounded-state models on 24 of 29 text benchmarks (Appendix~\ref{sec:full-results}), and its AIME score lies in the range of leading compact growing-history models (Section~\ref{sec:peer-capability}). The bounded-state baselines differ from EdgeRock in training data and post-training as well as in architecture, so this comparison shows where the class stands rather than the cost of constant memory alone. That cost is measured against EdgeRock's own source model, Qwen3.5-4B (Section~\ref{sec:source-retention}).

\subsection{A clear lead over 16 constant-memory models}
\label{sec:bounded-capability}
\begin{figure}[!htbp]
\centering
\includegraphics[width=\linewidth]{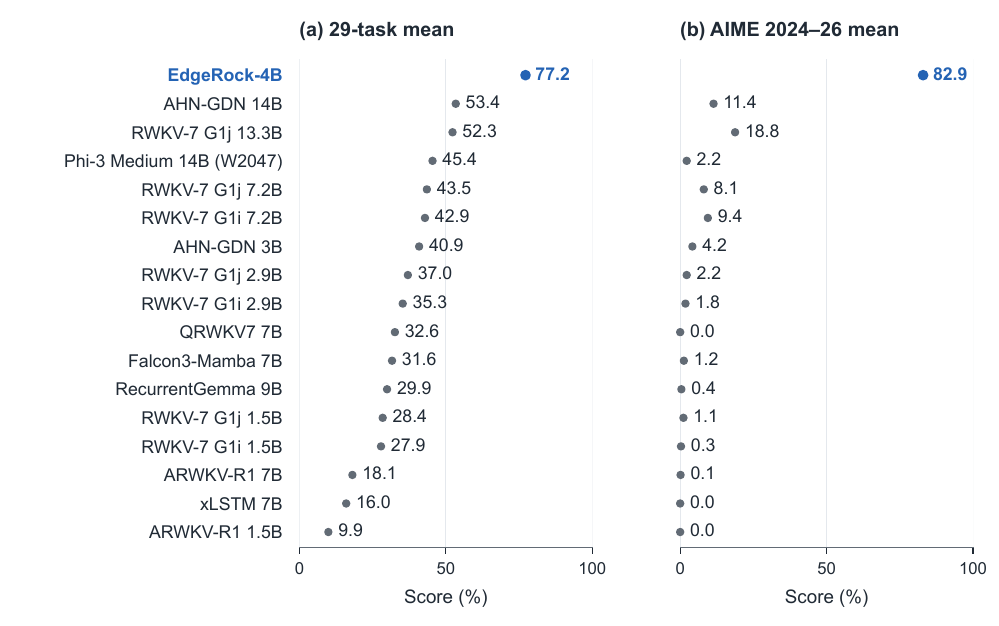}
\caption{\textbf{Bounded-state capability.} \textbf{(a)} Equal-weight mean across 29 text benchmarks. \textbf{(b)} Equal-weight mean across AIME 2024--2026. Both panels order models by the 29-benchmark mean. EdgeRock-4B ranks first in both summaries among 17 models spanning 1.5B--14B. W2047 identifies Phi-3 Medium's configured 2,047-token attention window.}
\label{fig:bounded}
\end{figure}

EdgeRock-4B leads the evaluated bounded-state class in both the 29-benchmark mean and competition mathematics (Figure~\ref{fig:bounded}). The comparison includes larger GDN, RWKV, and Mamba models~\cite{ahn,rwkvg1,falcon3}. Its equal-weight 29-benchmark mean exceeds AHN-GDN Qwen2.5 14B, the strongest baseline on this aggregate, by \textbf{23.8 percentage points}.

On AIME 2024--2026, EdgeRock-4B reaches an \textbf{82.9\%} mean pass@1, compared with 18.8\% for RWKV-7 G1j 13.3B, the strongest bounded-state AIME comparator (Figure~\ref{fig:bounded}).

\label{app:bounded}

\subsection{Retaining 96\% of the source model's reasoning}
\label{sec:preservation}
\label{sec:source-retention}
EdgeRock-4B starts from Qwen3.5-4B's pretrained weights, so this comparison is the closest controlled measure of what constant memory costs. The cost is small: EdgeRock-4B retains \textbf{96.3\%} of Qwen3.5-4B's equal-weight AIME 2024--2026 mean pass@1 after replacing full attention with local attention and a fixed-size anchor bank. It also reaches \textbf{97.5\%} on MATH-500 (Table~\ref{tab:core-reasoning}).

\CoreReasoningResults

\begin{figure}[!htbp]
\centering
\includegraphics[width=\linewidth]{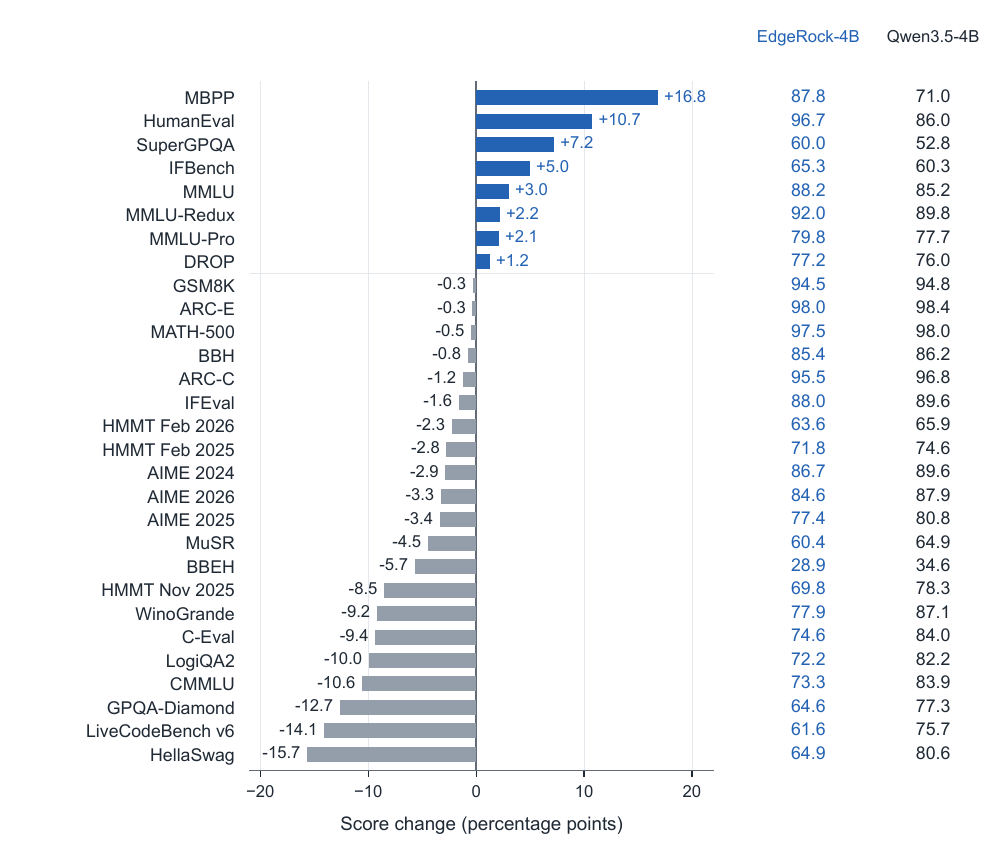}
\caption{\textbf{Capability after conversion.} Bars report EdgeRock-4B minus its Qwen3.5-4B source backbone in percentage points, ordered from largest to smallest. Blue highlights the eight benchmark gains; aligned columns give absolute scores (\%).}
\label{fig:delta}
\end{figure}

\Needspace{6\baselineskip}
EdgeRock-4B exceeds Qwen3.5-4B on eight of 29 text benchmarks, including MMLU, MMLU-Pro, MMLU-Redux, SuperGPQA, HumanEval, and MBPP (Figure~\ref{fig:delta}).

\label{sec:delta}

\subsection{In the range of compact models with growing memory}
\label{sec:peer-capability}
Against compact models whose histories grow, EdgeRock-4B reasons in the same range. On AIME 2024--2026, its 82.9\% lies between Granite4.2-3B (82.2\%) and Qwen3.5-4B (86.1\%), with MiniCPM5-2B at 87.8\%, and well above Nemotron-3-Nano-4B (57.1\%), LFM2.5-2.6B (56.2\%), Gemma4-E4B (40.1\%), and Phi-4-mini-reasoning (38.9\%) (Table~\ref{tab:peer-capability}; Appendix~\ref{app:growing-results}). EdgeRock-4B's advantage over MiniCPM5-2B is larger on MMLU-Pro and SuperGPQA, which test knowledge-intensive reasoning (Table~\ref{tab:peer-capability}).

EdgeRock-4B also leads LFM2.5-2.6B on \textbf{28 of 29 benchmarks} (Appendix~\ref{app:growing-results}).

\PeerCapabilityResults

\Needspace{17\baselineskip}
\subsection{Vision retained at constant memory}
\label{sec:vision}

EdgeRock-4B extends its constant-memory design to text-and-image reasoning. Across six visual benchmarks covering perception, chart and diagram understanding, and visual question answering, it reaches an \textbf{84.9\%} equal-weight mean answer-level accuracy, compared with 89.1\% for Qwen3.5-4B, retaining 95.3\% (Table~\ref{tab:visual-results}).

\begin{table}[!htbp]
\centering
\caption{Visual answer-level accuracy (\%), including MME. Retained gives EdgeRock-4B as a percentage of Qwen3.5-4B. Appendix~\ref{app:protocol} gives standard MME points.}
\label{tab:visual-results}
\begin{tabularx}{\linewidth}{@{}Yrrr@{}}
\toprule
Benchmark & EdgeRock-4B & Qwen3.5-4B & Retained (\%)\\
\midrule
MMBench EN~\cite{mmbench} & 88.3 & 90.1 & 98.0 \\
MMBench CN~\cite{mmbench} & 89.0 & 91.4 & 97.4 \\
POPE~\cite{pope} & 85.4 & 88.8 & 96.2 \\
MME~\cite{mme} & 83.0 & 86.8 & 95.6 \\
ChartQA~\cite{chartqa} & 83.9 & 90.3 & 92.9 \\
AI2D~\cite{ai2d} & 80.0 & 87.0 & 92.0 \\
\midrule
Equal-weight mean & 84.9 & 89.1 & 95.3 \\
\bottomrule
\end{tabularx}
\end{table}

\FloatBarrier

\FloatBarrier
\section{Do distant answers depend on RMA?}
\label{sec:diagnostic}

\begin{figure}[!htbp]
\centering
\includegraphics[width=\linewidth]{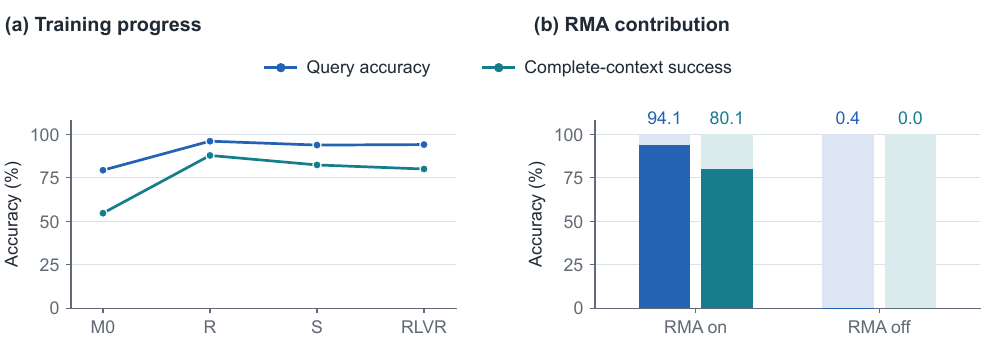}
\caption{\textbf{Distant-memory reasoning.} \textbf{(a)} Query accuracy and complete-context success after RMA adaptation (M0), joint adaptation (R), supervised fine-tuning (S), and reinforcement learning with verifiable rewards (RLVR). Complete-context success requires all four answers to be correct. \textbf{(b)} The final model with all eight RMA modules on or off; recurrence and local attention remain active in both settings. With RMA enabled, EdgeRock correctly answers 94.1\% of queries and completes 80.1\% of contexts across retrieval, state updates, cross-document combination, and variable binding.}
\label{fig:memory}
\end{figure}

RMA is what lets EdgeRock-4B use evidence beyond its local window. On 256 contexts of 16K to 130,000 input tokens, with four queries each covering retrieval, state updates, cross-document combination, and variable binding, it answers \textbf{94.1\%} of queries; with RMA disabled, it answers \textbf{0.4\%} (Figure~\ref{fig:memory}). Memory is learned early and kept through post-training: query accuracy rises from 79.4\% after RMA adaptation (M0) to 96.1\% after joint adaptation (R) and stays at 94.1\% in the final model (Figure~\ref{fig:memory}a).

\subsection{Without RMA, distant recall collapses}
\label{sec:memory-branch-intervention}
All eight RMA modules are disabled at inference time while the recurrent backbone and local attention remain active. Correct answers fall from 964 to 4 of 1,024 queries, and complete-context success from 205 to 0 of 256 contexts (Figure~\ref{fig:memory}). Because RMA is disabled only at inference, this shows that in the trained model, distant answers flow through RMA rather than through recurrence or the local window.

\subsection{RMA across memory tasks}
\paragraph{Task breakdown.}
EdgeRock-4B achieves 90.6\% complete-context success on exact retrieval, 87.5\% on state-update interpretation, and 89.1\% on cross-document combination (Table~\ref{tab:memory-task-breakdown}).

\paragraph{Paired evidence interventions.}
EdgeRock-4B answers both versions correctly in 89.8\% of 256 pairs where changed distant evidence requires a changed answer, and in 96.5\% of 256 pairs where the answer should stay the same (Table~\ref{tab:paired-memory-interventions}).

\begin{table}[!htbp]
\centering\small
\caption{Memory diagnostics. Top: complete-context success by task, requiring all four answers in a context to be correct (64 contexts per task). Bottom: paired changes to distant evidence, requiring both answers to be correct (256 pairs each); among pairs whose answer should stay the same, the answer never changed.}
\label{tab:memory-task-breakdown}\label{tab:paired-memory-interventions}
\begin{tabularx}{\linewidth}{@{}Yrr@{}}
\toprule
 & Correct / total & Rate (\%)\\
\midrule
\multicolumn{3}{@{}l}{\textit{Complete-context success by task}}\\
Exact retrieval & 58 / 64 & 90.6\\
State-update interpretation & 56 / 64 & 87.5\\
Cross-document combination & 57 / 64 & 89.1\\
Variable binding & 34 / 64 & 53.1\\
All tasks & 205 / 256 & 80.1\\
\addlinespace[4pt]
\multicolumn{3}{@{}l}{\textit{Paired changes to distant evidence}}\\
Answer should change & 230 / 256 & 89.8\\
Answer should stay the same & 247 / 256 & 96.5\\
\bottomrule
\end{tabularx}
\end{table}

\FloatBarrier
\section{What does constant memory buy in deployment?}
\label{sec:systems}

Constant memory turns directly into serving gains. On one Hopper 80GB GPU, EdgeRock-4B decodes a single request at \textbf{\EdgeLongTPS{} tokens/s} through 128K output, runs \textbf{512} long-generation sequences at once, and averages \textbf{\EdgeBatchRatio{}$\times$} Qwen3.5-4B's end-to-end throughput. The gains come from fixed history: per-token work does not grow with context, and one GPU runs more sequences at once. Implementation details such as CUDA Graphs and fused BF16 kernels are in Appendix~\ref{app:runtime}.

\FloatBarrier
\subsection{Steady decoding and higher throughput}
\label{sec:throughput}\label{sec:batch}

\begin{figure}[!htbp]
\centering
\includegraphics[width=\linewidth]{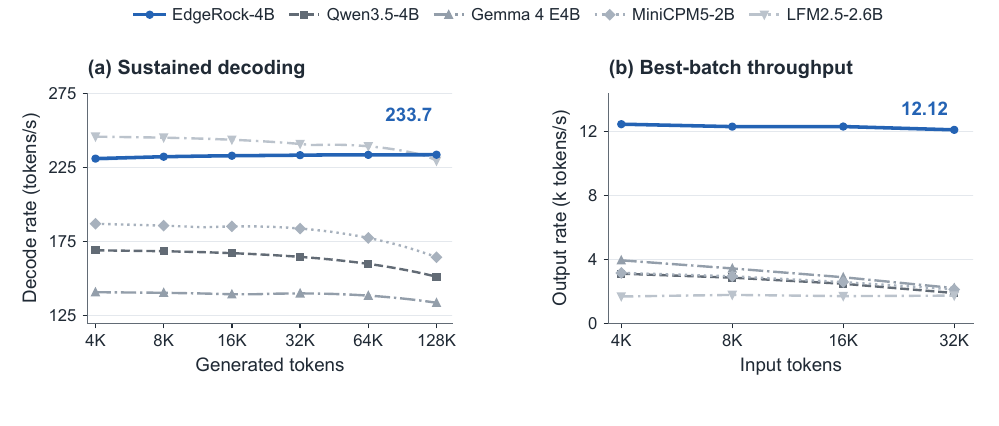}
\caption{\textbf{Long-context inference.} EdgeRock-4B and four compact models use BF16 weights and KV on one Hopper 80GB GPU. \textbf{(a)} Cumulative decode rate during 128K output after 128K input at batch size $1$. EdgeRock sustains \EdgeLongTPS{} tokens/s through the full output. \textbf{(b)} Output-token throughput for 4K--32K inputs and 32K output with repeated prompts. Only output tokens are counted; elapsed time includes prefill. Each model uses its best measured batch at each input length (Appendix Table~\ref{tab:peer-throughput}); EdgeRock leads across all four workloads.}
\label{fig:throughput}\label{fig:decode-speed}
\end{figure}

\paragraph{Single-request decoding.}
EdgeRock-4B sustains \textbf{\EdgeLongTPS{} tokens/s} through 128K output after 128K input, \textbf{\EdgeSingleRatio{}$\times$} Qwen3.5-4B's rate under the same workload at batch size $1$. Decoding remains nearly flat throughout generation (Figure~\ref{fig:decode-speed}a). With 254K input and 2K output, reaching the same 256K total length, EdgeRock-4B decodes at \textbf{226.6 tokens/s} versus Qwen3.5-4B's 133.9, a \textbf{1.69$\times$} gain.

\paragraph{Long-generation throughput.}
Across 4K, 8K, 16K, and 32K inputs with 32K output, EdgeRock-4B averages \textbf{\EdgeBatchRatio{}$\times$} Qwen3.5-4B's end-to-end throughput and \textbf{\EdgeLFMRatio{}$\times$} LFM2.5-2.6B's at each runtime's best measured batch (Appendix Table~\ref{tab:peer-throughput}). The gain comes from capacity. At 32K input and 32K output, each Qwen3.5-4B sequence ends with 2\,GiB of attention KV (32\,KiB per token), so its cache fills after a few dozen concurrent sequences on one 80GB GPU and further requests wait. EdgeRock's batch runtime keeps persistent history at 97.76\,MiB per sequence, so all 512 sequences remain active through the full output.

\FloatBarrier
\subsection{Concurrency scales to 512 sequences on one GPU}
\label{app:latency-memory}

\begin{figure}[!htbp]
\centering
\includegraphics[width=\linewidth]{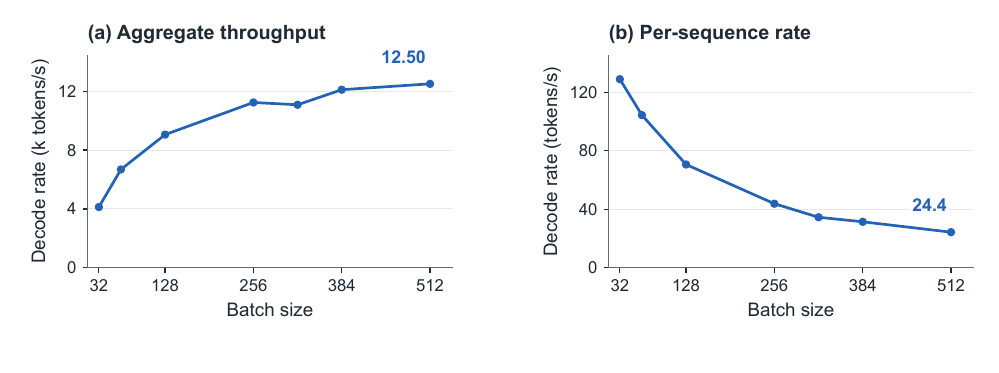}
\caption{\textbf{Concurrency scaling.} EdgeRock-4B with 32--512 active sequences on one Hopper 80GB GPU (BF16 weights and KV). Rates pool decode tokens and time across all 16 input/output pairs from 4K to 32K. \textbf{(a)} Aggregate throughput. \textbf{(b)} Mean rate per active sequence.}
\label{fig:batch-scaling}
\end{figure}

EdgeRock-4B supports \textbf{512 concurrent sequences} across all 16 input/output combinations of 4K, 8K, 16K, and 32K tokens on one Hopper 80GB. At batch size $512$, weighted decoding reaches \textbf{\EdgeWeightedTPS{} tokens/s}, or \textbf{\EdgePerSequenceTPS{} tokens/s per active sequence} (Figure~\ref{fig:batch-scaling}).

\FloatBarrier
\subsection{Efficient decoding at short contexts}
\label{sec:short}

EdgeRock-4B reaches a decode rate of \textbf{\EdgeShortTPS{} tokens/s} at batch size $1$ and \textbf{\EdgeShortBatchTPS{} tokens/s} at batch size $512$ for 4K input and 2K output, excluding prefill. At batch size $1$, peak tensor allocation is \textbf{\EdgeShortMemory{}\,GiB} (Appendix Table~\ref{tab:short-scaling}).

\FloatBarrier
\subsection{Memory known before reasoning starts}
\label{sec:cache}

\begin{figure}[!htbp]
\centering
\includegraphics[width=\linewidth]{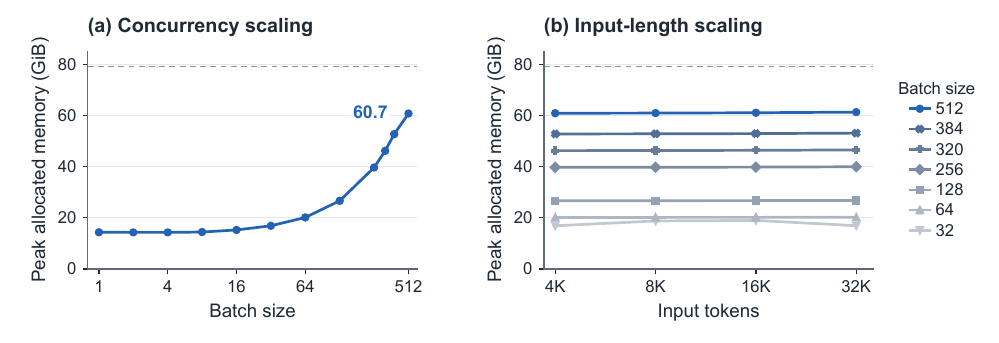}
\caption{\textbf{GPU memory scaling.} Peak tensor allocation of EdgeRock-4B on one Hopper 80GB GPU, including weights, history, buffers, and workspace; the dashed line marks GPU capacity. \textbf{(a)} By batch size, at 4K input and 2K output. \textbf{(b)} By input length, with 32K output; allocation stays nearly flat at each batch size.}
\label{fig:memory-capacity}
\end{figure}

In the batch runtime, each sequence carries \textbf{97.76\,MiB} of history, whether EdgeRock-4B reasons for 4K tokens or 256K (Table~\ref{tab:state-budget}). Its \textbf{72\,MiB} of attention KV is \textbf{99.1\%} smaller than Qwen3.5-4B's at 256K (Figure~\ref{fig:overview}b). On one Hopper 80GB GPU, a single 128K-input/128K-output request peaks at \textbf{\EdgeLongMemory{}\,GiB}, and 512 concurrent sequences need at most \textbf{\EdgeBatchMemory{}\,GiB} across all 16 workloads (Figure~\ref{fig:memory-capacity}).

\FloatBarrier
\Needspace{10\baselineskip}
\section{Conclusion}
\label{sec:conclusion}

EdgeRock shows that strong reasoning does not require a memory that grows with every token. Once the history a model carries is fixed, memory stops being a property of the input and becomes a property of the model: it can be sized once, at design time, and relied on for as long as the model reasons.

This changes what can be built around a reasoning model. A robot can run perception and reasoning in one continuous loop without its memory budget drifting as the task goes on. A dedicated chip can hold a model's entire history in on-chip memory, because that history no longer depends on context length. An agent's memory no longer grows with each tool call it makes. Serving systems can plan capacity by counting requests rather than by guessing how long each will run.

Several directions follow. Scaling RMA to larger backbones, beginning with additional EdgeRock sizes now in preparation, will test whether constant memory keeps pace with stronger reasoning. Applying the same conversion to other hybrid architectures would let existing models adopt constant memory without pretraining from scratch. Anchors could extend beyond text to streams of images, audio, and sensor data, where a growing history is most costly. And because each anchor records its source position, what a model chose to remember can be inspected directly.

\begingroup
\small
\urlstyle{same}
\setlength{\parskip}{0pt}

\endgroup

\clearpage
\appendix
\section{Evaluation and verification protocol}
\label{app:protocol}

\paragraph{Text evaluation and sampling}
Text evaluation uses fixed prompts, reference answers, and executable tests across 29 benchmarks.

\begin{table}[!htbp]
\small
\begin{tabularx}{\textwidth}{@{}p{4.7cm}X@{}}
\toprule
Parameter & EdgeRock-4B text evaluation setting\\
\midrule
Independent responses per item & $K_{\mathrm{eval}}=64$ for fewer than 100 questions; $16$ for 100--10,000 questions; $1$ above 10,000 questions\\
Sampling & Temperature 0.6; top-$p$ 0.95; top-$k$ 20; presence penalty 0\\
Seed / response cap & 20260923 / 131,072 tokens\\
Stopping & EOS, token cap, or an exact 128--256-token mechanical loop repeated eight times\\
\bottomrule
\end{tabularx}
\end{table}

Binary task scores report mean pass@1 with benchmark-prescribed subject or subtask aggregation. Repetition counts vary by model and benchmark. AIME pools responses within each edition and weights the 2024--2026 editions equally.

\paragraph{Task-specific verification and aggregation}
The verifier extracts a committed answer independently of the reference, preferring the final-answer region and accepting a terminal answer in the reasoning region for semantic scoring. Conflicting or unresolved answers receive no credit. Code scores use relaxed program extraction followed by executable tests. IFEval and IFBench use strict constraint checkers, with language-detector seed 0.

BBH, BBEH and CMMLU retain their subtask or subject macro-averages; MMLU-Redux uses corrected choices. MuSR averages its three task accuracies. DROP reports token F1 and exact match, with F1 contributing to the 29-benchmark mean. LogiQA2 uses accuracy on the supplied English panel. Benchmark means assign equal weight to each benchmark, retaining its designated primary metric.

\paragraph{Visual and long-memory evaluation}
\begin{table}[!htbp]
\small
\begin{tabularx}{\textwidth}{@{}p{3.1cm}X X@{}}
\toprule
Parameter & Visual evaluation & Long-memory diagnostics\\
\midrule
Generation & $K_{\mathrm{eval}}=1$; thinking enabled & Greedy $K_{\mathrm{eval}}=1$\\
Arithmetic / batch & Reference / 128 & Reference / 1\\
Sampling & Temperature 1.0; top-$p$ 0.95; top-$k$ 20; presence penalty 1.5 & Greedy selection\\
Seed & 20260912 & 2026091707\\
Response cap & 32,768 tokens & 1,024 tokens\\
Input extent & 65,536--4,194,304 pixels & Approximately 16K--130,000 tokens\\
\bottomrule
\end{tabularx}
\end{table}

The six reported visual benchmarks are AI2D, ChartQA, MMBench EN/CN, MME and POPE; the visual benchmark mean weights them equally. Answer-weighted accuracy, MME category points and MMBench circular accuracy remain separate metrics. EdgeRock scores 1,600.4 MME Perception points, 728.2 Cognition points, and 2,328.7 total points. MMBench EN circular accuracy is 81.6\%; six-benchmark answer-weighted accuracy is 85.5\%.

The memory suite has 256 contexts and 1,024 queries: 64 contexts each for exact retrieval, state-update interpretation, cross-document combination and variable binding, with four queries per context. Query accuracy measures individual answers; complete-context success requires all four answers to be correct. Paired interventions change distant evidence while holding the surrounding structure fixed, testing answers that should change or remain stable; paired accuracy requires \emph{both} answers to be correct. The branch-disabled intervention retains the trained backbone and local window.

\clearpage
\section{Complete bounded-state benchmark results}
\label{sec:full-results}
Results cover EdgeRock-4B and 16 bounded-state comparators on 29 benchmarks. Values use each task's primary metric on a 0--100 scale; bold marks column maxima, including ties. Appendix~\ref{app:growing-results} reports the growing-history models.

\BoundedMathResults
\BoundedKnowledgeResults
\FloatBarrier
\BoundedReasoningResults
\BoundedCodeInstructionResults
\FloatBarrier
\BoundedAdditionalResults
\FloatBarrier

\FloatBarrier
\Needspace{12\baselineskip}
\section{Complete growing-history benchmark results}
\label{app:growing-results}
Results compare EdgeRock-4B with eleven growing-history models on 29 benchmarks. Bold marks column maxima, including ties. Metrics and sampling follow Appendix~\ref{app:protocol}.

\GrowingMathResults
\GrowingKnowledgeResults
\FloatBarrier
\GrowingReasoningResults
\GrowingCodeInstructionResults
\FloatBarrier
\GrowingAdditionalResults
\FloatBarrier

\FloatBarrier
\section{Runtime protocols and memory accounting}
\label{app:runtime}

\subsection{Workloads and hardware}
All runs use one NVIDIA Hopper 80GB GPU and tensor-parallel size one. All matrix weights and KV are BF16.

Here $1\mathrm{K}=1{,}024$ tokens. The sustained single-request workload (batch size $=1$) is 128K input plus 128K output. A second single-request workload uses 254K input and 2K output, with three independent runs per model for EdgeRock and Qwen; both workloads total 262,144 tokens. Long-generation throughput uses 4K/8K/16K/32K inputs and 32K output. The full EdgeRock sweep combines those four input lengths with the same four output lengths at batches 32, 64, 128, 256, 320, 384, and 512. A separate 4K-input/2K-output sweep covers batch sizes 1, 2, 4, 8, 16, 32, 64, 128, 256, 320, 384, and 512. Each EdgeRock configuration uses three independent runs; rates are medians unless specified as weighted.

The benchmark uses synthetic input tokens, EOS ignored, and a 256-token warmup at the target input length and batch capacity. EdgeRock uses full-vocabulary autoregressive sampling with temperature 1.0, top-$k$ 20, top-$p$ 0.95, and presence penalty 1.5.

\subsection{Timed intervals and aggregation}
\textbf{Decode rate} measures output after each request's first token, including sampling, state updates, graph management, and runtime scheduling. The single-request curves report cumulative rate every 4K output tokens. \textbf{End-to-end output-token throughput} divides the total number of generated output tokens by the full generation-call duration, including prefill time. Input tokens do not enter the token count. Both exclude model loading, warmup, tokenization, and text decoding.

Mean throughput ratios weight the four 32K-output workloads equally, using each runtime's best measured batch. Weighted batch throughput divides total subsequent output tokens by total decode time across all 16 length pairs and three runs; per-sequence rate divides it by the active batch. Bands show the range across runs. Memory results use peak allocated and reserved memory across runs.

\subsection{Runtime configuration and prompt reuse}
Packed projections and fused normalization, rotary encoding, and memory operations reduce data movement. At batch size $512$, the runtime shares original and packed BF16 weight storage, saving 4.95\,GiB. At batch size $1$, the runtime fuses output-head processing with full-vocabulary sampling and state updates. Decoding is autoregressive, with weights and history on the GPU.

\begin{table}[!htbp]
\centering\small
\caption{Runtime configuration of each model on one Hopper 80GB GPU.}
\label{tab:runtime-configurations}
\begin{tabularx}{\linewidth}{@{}>{\raggedright\arraybackslash}p{.20\linewidth}>{\raggedright\arraybackslash}p{.23\linewidth}Y@{}}
\toprule
Model & History precision & Execution path\\
\midrule
EdgeRock-4B & BF16 KV; GDN: FP32 at batch size $1$, FP16 otherwise; FP32 updates & Custom CUDA Graph engine; fused BF16 kernels; directly indexed state and local rings\\
\addlinespace[4pt]
Qwen3.5-4B & BF16 KV; FP32 GDN state & vLLM~\cite{pagedattention} CUDA Graphs; FlashAttention-3~\cite{flashattention3} (batch size $1$), -4~\cite{flashattention4} (batch); FlashInfer~\cite{flashinfer} GDN prefill and fused decode\\
\addlinespace[4pt]
Gemma 4 E4B & BF16 KV & vLLM O3; FlashAttention-4; language model only\\
\addlinespace[4pt]
MiniCPM5-2B & BF16 KV & vLLM O3; FlashAttention-3\\
\addlinespace[4pt]
LFM2.5-2.6B & BF16 KV & vLLM 0.23.0; CUDA Graphs and chunked prefill\\
\bottomrule
\end{tabularx}
\end{table}

EdgeRock and LFM test batches 32, 64, 128, 256, 320, 384, and 512; Qwen and MiniCPM test 32 and 64; Gemma tests 64 and 128. Batch denotes submitted requests. EdgeRock maintains the target concurrency; vLLM schedules active requests within its cache budget. Eight continuations share each prompt in the EdgeRock batch grid, then retain independent recurrent, local-attention, and anchor state. The short sweep shares a prompt across up to eight requests.

The vLLM peers use CUDA Graphs, chunked prefill, and a 16K prefill-token budget. Qwen, Gemma, and MiniCPM use asynchronous scheduling. Gemma, MiniCPM, and LFM use 95\% GPU-memory utilization. The synthetic single-request workloads use a 256K execution capacity with unchanged weights and RoPE frequencies. Gemma E4B denotes effective scale; its full checkpoint is approximately 8B parameters. MiniCPM's checkpoint is approximately 2.52B.

\subsection{History and device allocation}
The 72\,MiB of attention KV is a configuration-derived bound. During decoding, the local cache stores 1,023 preceding positions and the decoder supplies the current token. The history tensors in Table~\ref{tab:state-budget} total 97.76\,MiB per sequence in the batch runtime, including recurrent state, convolution state, and anchor metadata.

Peak tensor allocation includes weights, history, input/output buffers, graph storage, and temporary workspace. Reserved memory also includes cached allocator blocks. At batch size $512$, shared storage for original and packed BF16 projections eliminates duplicate weights while retaining the full history capacity.

\subsection{Measured throughput and memory}
Tables~\ref{tab:peer-throughput}--\ref{tab:short-scaling} list the measurements behind Figures~\ref{fig:throughput}--\ref{fig:memory-capacity}.

\PeerThroughputResults
\BatchScalingResults
\ShortScalingResults

\FloatBarrier
\section{Streaming execution and state ownership}
\label{sec:streaming-execution}

\subsection{State owned by a continuing request}

A continuing request owns recurrent matrices, a local KV ring, committed anchor banks, pending candidates, and progress metadata (Table~\ref{tab:streaming-state}). This state persists across pauses and slot transfers. New anchors become readable after the next block commit.

\begin{table}[!htbp]
\centering\small
\caption{Logical state needed to continue one request. Capacities are per RMA module and KV group where applicable.}
\label{tab:streaming-state}
\begin{tabularx}{\linewidth}{@{}p{.23\linewidth}YY@{}}
\toprule
State & Contents & Maintenance rule\\
\midrule
Recurrent state & The GDN history representation & Update fixed-dimensional recurrent matrices.\\
\addlinespace[3pt]
Local history & Recent attention KV and the ring position & Retain at most $W=1{,}024$ positions.\\
\addlinespace[3pt]
Committed anchor bank & Selected KV records, admission scores, and source positions & Retain at most $K=1{,}024$ records per group.\\
\addlinespace[3pt]
Pending candidates & Current-block records, scores, and occupancy & Accumulate up to $C=256$ positions before a commit.\\
\addlinespace[3pt]
Progress metadata & Position and block progress, plus association with the active request & Preserve consistent continuation and slot ownership.\\
\bottomrule
\end{tabularx}
\end{table}

Model weights are shared across requests; history occupies independent slots in shared cache pools. Extending a request reuses its history capacity, while admitting another request allocates an additional slot.

\Needspace{32\baselineskip}
\subsection{Forward procedure and causal visibility}

Prefill and decode follow Algorithm~\ref{alg:anchor-update} with the same logical commit boundaries. Queries read the committed bank throughout each 256-token block, while local attention covers recent tokens. A boundary commit makes selected candidates available to the next block.

\begin{algorithm}[!htbp]
\caption{Causal RMA update in one local-plus-RMA layer}
\label{alg:anchor-update}
\small
\begin{algorithmic}[1]
\Require Hidden input $h_t$, causal index $t$, rotary coordinates $\mathbf p_t$
\Require Local state $\mathcal L$, committed anchor bank $\mathcal M$, pending candidates $\mathcal C$
\Require Window $W=1{,}024$, bank capacity $K=1{,}024$, candidate block size $C=256$
\State $u_t\gets\mathrm{Norm}_{\mathrm{mem}}(h_t)$
\State $m_t\gets\mathrm{ReadAndGate}(u_t,\mathbf p_t,\mathcal M)$ \Comment{Committed bank only}
\State $w_t\gets\mathrm{Writer}(u_t)$; $(k_t,v_t)\gets\mathrm{MemoryKV}(w_t,\mathbf p_t)$
\State $a_t\gets\mathrm{Admission}(u_t)$ \Comment{Four causal group scores}
\State $(\ell_t,\mathcal L)\gets\mathrm{LocalAttentionStep}(h_t,\mathbf p_t,\mathcal L;W)$
\State $z_t\gets h_t+\ell_t+m_t$; $h_t^{\mathrm{out}}\gets z_t+\mathrm{MLP}(\mathrm{Norm}_{\mathrm{ffn}}(z_t))$
\State Append $(k_t,v_t,a_t,t)$ to $\mathcal C$
\If{$|\mathcal C|=C$} \Comment{Logical block boundary, independent of prefill chunks}
  \For{$g=1,\ldots,4$}
    \State $\mathcal M_g\gets\operatorname{TopK}_{a_g}(\mathcal M_g\cup\mathcal C_g;K)$
  \EndFor
  \State $\mathcal C\gets\varnothing$ \Comment{Updated bank is visible to the next block}
\EndIf
\State \Return $h_t^{\mathrm{out}},\mathcal L,\mathcal M,\mathcal C$
\end{algorithmic}
\end{algorithm}

\texttt{ReadAndGate} combines query projection, committed-bank attention with the null item, the signed gate, and output projection (Section~\ref{sec:memory-algorithm}). \texttt{MemoryKV} applies the writer's key and value projections, key normalization, and rotary encoding. Local attention includes the current token and retains at most $W=1{,}024$ positions. Top-$K$ ties favor earlier source positions. A new sequence starts with empty local state, anchor banks, and candidate buffers; unfinished candidate blocks persist across prefill chunks and request pauses.

\subsection{Selection invariants and information availability}

Each KV group examines at most 1,280 records at a boundary and retains up to 1,024, preserving the surviving key and value vectors unchanged. Fixed write-time scores and consistent tie ordering maintain the history top-$K$ invariant in Equation~\ref{eq:history-topk}. Query heads share the group's bank and compute independent retrieval weights. Local and memory attention normalize separately; an empty bank returns zero.

\subsection{Continuation, chunking, and request lifecycle}

The portable backend defaults to 1,024-token prefill chunks. Chunking bounds the working set while preserving local, recurrent, and anchor state. RMA commits remain tied to completed candidate blocks, independently of execution chunk boundaries.

Prefix resumption restores every component in Table~\ref{tab:streaming-state}, including unfinished candidates, so their records participate in the next commit.

Native adapters access history in place through GPU request-slot indices. SGLang~\cite{sglang} reset and copy operations initialize reassigned slots and preserve continuing requests.

\paragraph{Released positional configuration.}
RMA uses partial RoPE over 64 of 256 coordinates, base $10^7$, default scaling, and a maximum-position execution setting of 262,144. Multimodal positions use interleaved rotary sections $(11,11,10)$; text supplies the same cumulative token index on all three axes~\cite{yanchorcode,yanchorconfig}.

\end{document}